%% file: main.tex
\documentclass[11pt]{article}

\usepackage[preprint]{acl}

\usepackage{times}
\usepackage{latexsym}
\usepackage[T1]{fontenc}
\usepackage[utf8]{inputenc}
\usepackage{microtype}
\usepackage{inconsolata}
\usepackage{graphicx}
\usepackage{booktabs}
\usepackage{array}
\usepackage{multirow}
\usepackage{arydshln}
\usepackage{xurl}  

\graphicspath{{figures/}}

\newcommand{\ci}[2]{{\scriptsize[#1, #2]}}

\title{Controlling Vocabulary, Syntax, and Length in Image Descriptions}

\author{Naoto Nishida and Yoshio Ishiguro \\ The University of Tokyo \\ \texttt{nawta@g.ecc.u-tokyo.ac.jp}}

\begin{document}
\maketitle

\input{00_abstract}
\input{00_architecture}
\input{01_introduction}
\input{02_related_work}
\input{03_method}
\input{05_results}
\input{05_cross_axis}
\input{07_human_evaluation}
\input{07b_human_quality}
\input{07_followup}
\input{07_discussion}
\input{06_conclusion}
\input{08_limitations}

\bibliography{references}

\clearpage
\appendix
\input{09_appendix}
\input{10_additional_results}
\input{11_transfer_rules}
\input{12_cross_axis_details}
\input{13_human_evaluation_details}

\end{document}

%% file: 00_abstract.tex
\begin{abstract}
    Adapting text to readers' knowledge and purposes requires identifying linguistic demands and generating text that meets them.
    We study the generation step, taking vocabulary, syntax, and length targets as given: an image description may need to retain necessary terms while using short, simple sentences.
    Motivated by future on-device use, we distill a 125B teacher's generated descriptions into Qwen3-VL vision-language models (VLMs) with 4B, 8B, and 32B parameters.
    We filter teacher outputs by linguistic rules and train students on retained text using supervised fine-tuning (SFT), then apply direct preference optimization (DPO) to pairs generated by each student.
    We compare these training stages with prompting across eight target combinations.
    Automatic evaluation uses two sets of 100 images excluded from task-specific training, plus a third set of 100 images frozen before generation; our primary metric requires the exact requested word ranges.
    Averaged across the first two sets, 4B strict joint success rises from 21.8~\% with prompting to 47.7~\% with SFT+DPO.
    Students at all sizes improve over prompting on both image sets, and a blinded human study finds no detectable loss in rated faithfulness or naturalness after training.
    These results show that training on filtered teacher-generated text, followed by preference optimization, improves how often an image description meets vocabulary, syntax, and length requirements together.
\end{abstract}

%% file: 00_architecture.tex
\begin{figure*}[t]
  \centering
  \includegraphics[width=\textwidth,pagebox=cropbox]{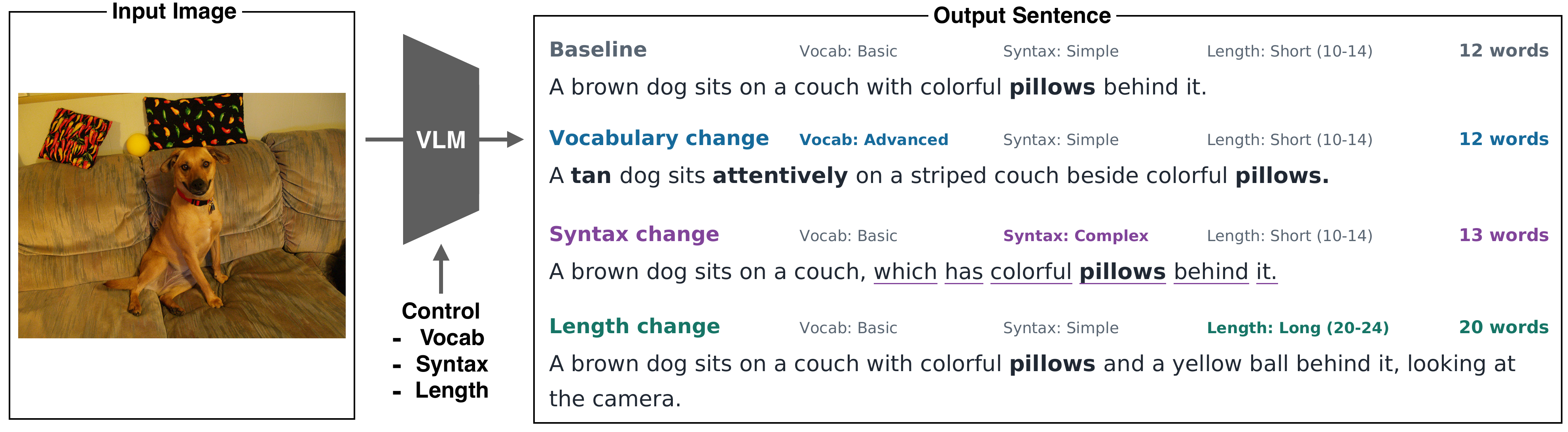}
  \caption{Task overview: a vision-language model receives an image and vocabulary, syntax, and length instructions to generate a description. These illustrative outputs come from the supplementary 32B model trained on 4B-generated targets with SFT and DPO (Appendix~\ref{app:small-teacher-method}). ``Baseline'' denotes the basic/simple/short request; each other row changes one requested attribute. Within sentences, bold marks lexicon-matched B1-or-higher words; underlining marks a detected finite relative clause. All four outputs meet the strict rules; Section~\ref{sec:task} defines these rules and vocabulary bands. Photo: COCO 347930 (CC BY 2.0 per dataset metadata).}
  \label{fig:architecture}
\end{figure*}

%% file: 01_introduction.tex
\section{Introduction}

Adapting text to readers' knowledge and purposes helps them understand and use information.
U.S. plain-language guidance asks writers to consider their audience's knowledge, expertise, and interests \citep{gsa-plainlanguage}.
In a study of scientific summaries, \citet{august-etal-2024-know} found that simpler text gave readers with little topic familiarity a better reading experience, while more familiar readers tended to overlook details such as study limitations in overly plain summaries.
A single version for a ``general audience'' may therefore serve readers with different backgrounds unevenly.

Adapting generated text involves identifying suitable linguistic demands for a reader and producing text that meets them.
We study image-description generation under supplied vocabulary, syntax, and length targets, for example from an assessment or settings chosen by the user.

This generation step may require separate targets for different linguistic properties.
Health communication guidance recommends short, simple sentences while retaining necessary technical terms with plain-language explanations \citep{odphp-2025-plain}.
In language learning, \citet{jin-etal-2026-toward} show that controlling large language model (LLM) outputs can improve comprehensibility for beginners.
An exercise introducing unfamiliar words may need those words within familiar syntax and short sentences, whereas an exercise introducing a new clause structure may need familiar vocabulary.
Meeting such requests requires controlling vocabulary, syntax, and length separately while maintaining the other requirements when one target changes.

Controllable text simplification addresses several of these properties.
ACCESS controls length, paraphrasing, lexical complexity, and syntactic complexity when rewriting a source sentence \citep{martin-etal-2020-controllable}.
With an image as input, the model selects what to describe and expresses it under the requested linguistic constraints, without a source sentence to rewrite.
Vision-language models (VLMs) generate descriptions from images and instructions; a useful description needs to represent the scene accurately and express it in language accessible to its reader.
A non-native speaker may recognize a photographed scene yet struggle with unfamiliar words or complex syntax in its description.
Photograph-description activities also appear in Cambridge English learning materials \citep{cambridge-b1-speaking}.
Example sentences supporting these activities need to match both the image and the linguistic demands of the exercise.
Figure~\ref{fig:architecture} illustrates different requests for the same scene.

For uses involving personal photographs, generating descriptions locally would avoid sending images to a remote service.
This prospect motivates our focus on a small VLM, with larger models as comparisons.
We ask whether task-specific training improves adherence to linguistic instructions and how the gains vary with model size.
We use a 125B teacher to generate descriptions and retain those that pass the linguistic rules and an automatic image-faithfulness screen.
The retained text provides supervision for training 4B, 8B, and 32B students.

Across eight combinations of vocabulary, syntax, and length targets, we compare prompting, supervised fine-tuning (SFT), and SFT followed by direct preference optimization (DPO) in 4B, 8B, and 32B students.
Training improves joint constraint compliance over prompting at all sizes.
Our evaluation distinguishes joint success from two effects of changing a request: moving the intended attribute and preserving the other requirements.
For the final 4B SFT+DPO model, we measure these effects and test how vocabulary and syntax requests correspond to human difficulty ratings, including effects on the other rating dimension.

%% file: 02_related_work.tex
\section{Related Work}

Controllable generation addresses several properties relevant to linguistic difficulty.
ACCESS controls lexical complexity, syntactic complexity, and length when rewriting a source sentence \citep{martin-etal-2020-controllable}.
In image captioning, constrained beam search enforces selected words \citep{anderson-etal-2017-guided}, while part-of-speech conditioning guides sentence structure \citep{Deshpande_2019_CVPR}.
\citet{deng-etal-2020-length} and CLID \citep{hirsch-tal-2024-clid} target caption length.
CaptionSmiths jointly controls length, descriptiveness, and word uniqueness, and analyzes how the other properties change when one condition varies \citep{saito-etal-2025-captionsmiths}.
Our task combines vocabulary targets defined by proficiency word lists, clause structure, and word-count ranges in descriptions generated directly from images.
We measure joint pass rates and test whether changing one request moves the intended attribute while the other two still meet their specified criteria.

Constraint satisfaction also connects our task to instruction-following evaluation.
IFEval distinguishes individual-instruction accuracy from satisfaction of all instructions in a prompt \citep{zhou2023instruction}, and FollowBench evaluates increasingly complex combinations of constraints \citep{jiang-etal-2024-followbench}.
NeuroStructural Decoding tracks lexical and dependency constraints during generation \citep{bastan-etal-2023-neurostructural}.
We use paired comparisons across eight combinations of linguistic targets, holding the image and two requests fixed within each pair.
Human ratings assess whether vocabulary and syntax requests correspond to perceived difficulty and how other requests affect those ratings.

Task-specific training can support these controls.
DPO learns from preferred and rejected responses \citep{rafailov-etal-2023-dpo}; prior work addresses its tendency to favor longer responses \citep{park-etal-2024-disentangling} and constructs preferences incorporating requested length \citep{yuan-etal-2025-following}.
Our preference pairs cover vocabulary, syntax, and length: preferred responses satisfy all strict rules, whereas rejected responses violate exactly one axis.
Comparing prompting, SFT on filtered teacher descriptions, and subsequent DPO at 4B, 8B, and 32B tests how training improves joint compliance across model sizes.

%% file: 03_method.tex
\section{Task and Evaluation}
\label{sec:task}

Given an image, a model must describe its main visible scene in one English sentence under three independently specified instructions: vocabulary (basic or advanced), syntax (simple or complex, labelled ``Grammar'' in the prompt), and length (10--14 or 20--24 words). Combining the two levels of each instruction gives eight conditions per image. These are operational control targets, not validated learner proficiency levels. In particular, a sentence labelled ``advanced'' need not correspond to any single Common European Framework of Reference (CEFR) level.

\paragraph{Automatic criteria.}
The CEFR-J and Octanove lexicons assign proficiency bands to words; we count words at B1 or above as higher-level vocabulary. Among recognized nouns, verbs, adjectives, and adverbs, basic requires at most 25\% higher-level words; advanced requires at least 35\% and at least two such words. Vocabulary fails when fewer than half the content words have lexicon entries. Grammar is simple when a dependency parser detects no qualifying relative or subordinate clause and complex when it detects at least one. The detector primarily identifies finite clauses, with an additional relative-pronoun heuristic. Length is scored against the exact requested range; a secondary variant allows a two-word tolerance (8--16 words for short and 18--26 for long). A format check requires a completed, single-sentence response without headings or other prefatory text.

Our primary outcome, \emph{strict joint success}, is the fraction of outputs satisfying vocabulary, grammar, the exact requested length range, and format together. We also report a secondary variant that allows a two-word tolerance around the requested length range (8--16 words for short and 18--26 for long requests); Table~\ref{tab:main} uses this tolerance variant, and Table~\ref{tab:strict} reports the primary metric. We also report each axis separately. For differences in rates or word shares, \% denotes subtraction of the two percentages. Faithfulness and naturalness are requested but are not part of joint success. Appendix~\ref{app:rules} provides exact rules. Training-example and SFT-checkpoint selection use the same vocabulary, grammar, and format checks, with strict length ranges.

We introduced the two-word tolerance after inspecting strict results. Because a tolerance chosen after seeing outcomes can flatter the methods, we treat exact adherence as primary and retain the tolerance variant as a sensitivity analysis (Appendix~\ref{app:additional}).

\paragraph{Images and uncertainty.}
We use COCO 2017 validation images \citep{lin-etal-2014-coco}, split by image into 500 training, 50 development, and 100 evaluation images (set 1) after duplicate filtering. A further 100-image split (set 2) was sampled after inspecting initial results. Each evaluation set contains 800 outputs per method. We additionally froze a confirmation set of 100 COCO images that appear in none of the earlier splits: checkpoints, prompts, and the strict scoring protocol were fixed before any confirmation outputs were generated, and we use this set for the confirmatory comparisons in Section~\ref{sec:results}. A 20-image development pilot fixed the prompts and rules; the full development split selected SFT checkpoints. Set 1 had already been inspected in earlier experiments with 4B-generated training descriptions (Appendix~\ref{sec:teacher}) and a 125B prompt reference before we designed the 125B-teacher experiment. Results on set 2 have also been inspected. We therefore treat both as exploratory evaluations. We report 95\% percentile bootstrap intervals using 10,000 image-level resamples; paired method comparisons retain all eight conditions and both methods for each sampled image. These intervals describe image-sampling uncertainty, not variation across training runs.

\section{Training and Comparisons}
\label{sec:method}

\begin{figure*}[t]
  \centering
  \includegraphics[width=\textwidth]{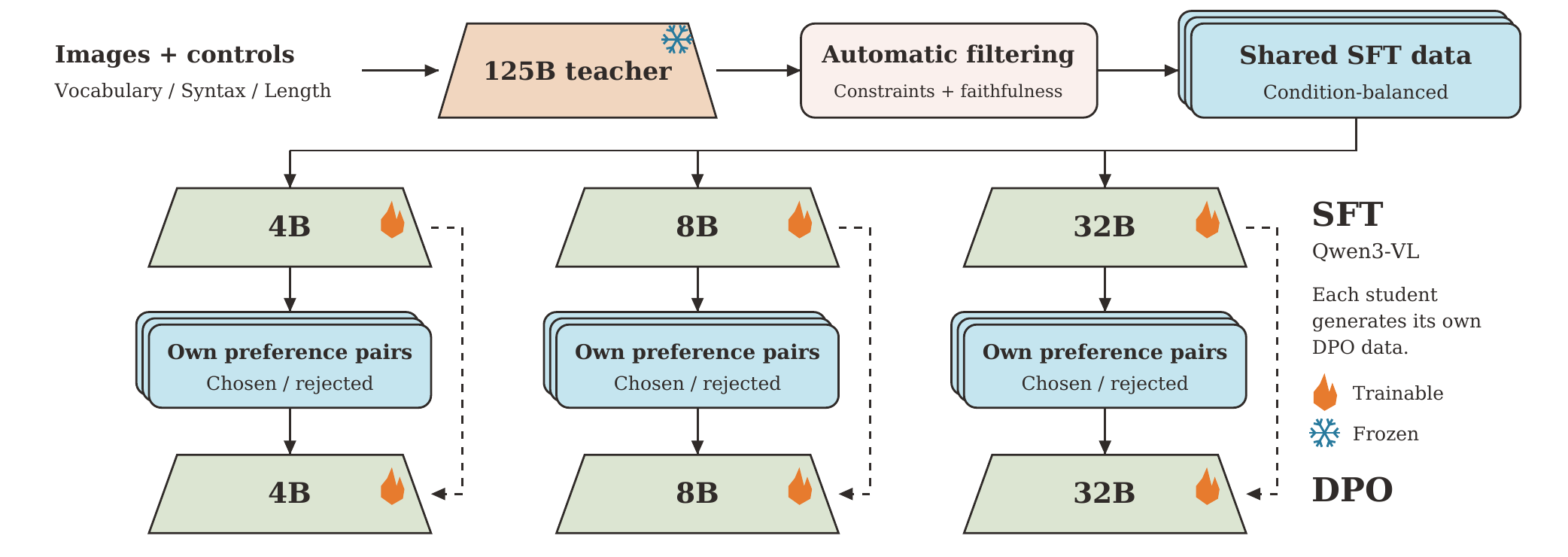}
  \caption{Training procedure. Filtered descriptions from the frozen 125B teacher form a shared dataset for supervised fine-tuning (SFT) of 4B, 8B, and 32B students. Each student then generates its own preferred/rejected pairs for direct preference optimization (DPO). Dashed arrows carry the SFT weights into DPO; flames mark trainable adapters. Data selection retains equal counts per SFT condition and per violated DPO axis; pair counts for simple and complex requests can differ. Faithfulness filtering uses an automatic model-based screen.}
  \label{fig:training-stages}
\end{figure*}

We train Qwen3-VL-Instruct students with 4B, 8B, and 32B parameters \citep{qwen3-vl} by supervised fine-tuning on 125B-teacher sentences, followed by DPO using rule-labelled pairs from each student's own outputs (Figure~\ref{fig:training-stages}). We compare each student with its unadapted model using the same images, eight instructions, and single greedy completion per condition. Direct student generation uses bfloat16 weights and a 96-token generation limit. Appendix~\ref{app:training} records model revisions and the hardware differences between runs.

\paragraph{Supervised fine-tuning (SFT).}
Training on complete teacher-generated descriptions follows sequence-level distillation \citep{kim-rush-2016-sequence}.
The teacher is Qwen3.8-Flash-Next, a 125B mixture-of-experts model served through llama.cpp with Q5\_K\_M quantization. It samples four descriptions for each training image and condition, followed by four more where none passes the strict rules. We retain candidates that pass those rules, a repetition filter, and a separate faithfulness screen from the unadapted 4B student model. Selecting at most one sentence per image--condition pair and retaining equal condition counts gives 1,752 examples (219 per condition). These targets have not been human-verified. An independent audit of the faithfulness screen on all 5,828 rule-passing candidates (Appendix~\ref{app:screen-audit}) shows that the screen is conservative: it removes 4.2\% of the candidates and retains 95.8\%, the share of image-contradicted sentences among the removed candidates (30.2\%) is 3.5 times that among retained candidates (8.5\%), and 46.5\% of the removed sentences were nevertheless judged supported. All three students receive the same examples and train LoRA adapters \citep{hu-etal-2022-lora} for three epochs. We select each student's epoch with highest development joint success under strict length ranges. The teacher's greedy descriptions provide an additional reference, with differences in model family, quantization, and inference backend described in Appendix~\ref{app:training}.

\paragraph{Direct preference optimization (DPO).}
Each selected SFT student generates its own candidates on the training images. We construct pairs sharing an image and prompt: the preferred response passes all strict rules, while the rejected response fails exactly one axis. Both pass the 4B faithfulness screen and repetition filter; we also exclude pairs with little overlap in content words to limit unrelated comparisons. Equal counts per violated axis yield 576, 462, and 426 pairs for the 4B, 8B, and 32B students, respectively. Each dataset has equal counts of basic and advanced vocabulary targets and of too-short and too-long length rejections; counts for simple and complex requests depend on the available candidates. DPO \citep{rafailov-etal-2023-dpo} uses one epoch, the student's selected SFT checkpoint as a frozen reference, and $\beta=0.1$. We denote the resulting method by SFT+DPO. Appendix~\ref{app:training} gives the overlap criterion and pair counts, and describes the earlier experiment using 4B-generated targets and shared preference pairs.

\paragraph{ACCESS comparison.}
Each unadapted Qwen model first generates one caption per image without linguistic targets.
The same pretrained ACCESS rewriter \citep{martin-etal-2020-controllable} then adapts that caption to each of the eight requests.
We select its controls on the 50 development images separately for each source-model size and request combination (Appendix~\ref{app:access}).

\paragraph{CaptionSmiths comparison.}
We train a 7B CaptionSmiths implementation \citep{saito-etal-2025-captionsmiths} on the 2,502 COCO reference captions for our 500 training images.
Training uses one warmup epoch followed by five epochs of condition training.
The students instead use teacher-generated descriptions of those images.
CaptionSmiths receives numeric length, descriptiveness, and word-uniqueness controls, selected on the 50 development images for each request combination.
We evaluate both its standard question and a variant that adds our task instruction, with separate control selection for each (Appendix~\ref{app:captionsmiths}).

%% file: 05_results.tex
\section{Results and Analysis}
\label{sec:results}

\begin{table*}[t]
  \centering
  \small
  \input{tables/teacher_main}
  \caption{Constraint pass rates (\%). Ours groups the direct-generation Prompt, SFT, and SFT+DPO comparisons. Students use filtered 125B-teacher descriptions for SFT and their own candidates for DPO. ACCESS rewrites captions from an unadapted Qwen model; Size denotes the Qwen source model, with the same ACCESS rewriter at all sizes. Set 1 and set 2 are separate evaluation sets of 100 images each. Joint success requires vocabulary, syntax, length, and format together. The three axis columns use set 1. Length allows two words outside the requested range; each cell uses 800 outputs (100 images $\times$ eight conditions). Brackets are 95\% image-bootstrap intervals. Within each Qwen size, bold marks the best displayed rate and underlining the second best; ties share a rank. ${}^\dagger$CaptionSmiths uses five epochs of condition training on COCO reference captions for the same training images; the second row adds our task instruction (Appendix~\ref{app:captionsmiths}).}
  \label{tab:main}
\end{table*}

\paragraph{Distillation improves constraint compliance.}
SFT on the teacher's filtered descriptions improves joint success over prompting at every student size on both image sets (Table~\ref{tab:main}).
With SFT alone, 4B success rises from 36.0\% to 59.4\% on the first set and from 35.4\% to 55.6\% on the second.
For 4B, SFT+DPO improves joint success over prompting by 24.5\% and 21.9\% on the two sets, respectively.
Each trained student also has higher vocabulary, syntax, and length pass rates than its prompt-only counterpart on both sets.
Appendix~\ref{app:additional} reports joint success by condition; Table~\ref{tab:axis-conditions2} gives second-set SFT+DPO axis rates by condition.

Under the primary strict metric (exact word ranges), the same ordering holds on both sets and on the frozen confirmation set (Table~\ref{tab:strict}).
On the confirmation set, SFT raises strict joint success from 24.5\% to 48.1\% at 4B, from 24.3\% to 49.1\% at 8B, and from 19.1\% to 50.8\% at 32B.
The paired gains over prompting are 23.6, 24.9, and 31.6 percentage points, respectively; Appendix~\ref{app:additional} reports the paired intervals.

\begin{table*}[t]
  \centering
  \small
  \input{tables/strict_main}
  \caption{Strict joint success (\%): vocabulary, syntax, the exact requested word range, and format together. Sets 1 and 2 are the original evaluation sets; the confirmation set was frozen, with checkpoints, prompts, and scoring fixed, before generation. Each cell uses 800 outputs. Within each size and column, bold marks the best rate and underlining the second best; ties share a rank.}
  \label{tab:strict}
\end{table*}

\paragraph{DPO gives smaller additional gains.}
SFT accounts for most of the improvement over prompting.
DPO raises the joint-success point estimate at every size on both sets, but its contribution varies by size.
At 8B, it adds 3.1\% on the first set and 6.9\% on the second.
At 4B and 32B, the additional gain is uncertain on both sets.
Appendix~\ref{app:training-differences} reports paired differences and confidence intervals.
On the confirmation set under strict scoring, DPO's additional gain over SFT is 0.6\% at 4B, 3.4\% at 8B (95\% CI [+0.1, +6.8]), and 1.8\% at 32B.
We therefore present SFT as the cost-efficient baseline and DPO as a refinement whose benefit depends on model size and instruction type; its clearest gains appear on the changed multi-sentence instructions of Section~\ref{sec:followup} (+7.0\% at 4B and +5.4\% at 8B).
For the 4B student, the DPO stage adds about 8.0 A100 GPU-hours on top of SFT's 5.51 (Appendix~\ref{app:training}).

\paragraph{Comparisons with prior methods.}
The ACCESS pipelines achieve 21.6--24.0\% joint success across source-model sizes and image sets.
At 4B, SFT+DPO exceeds this baseline by 36.5\% and 34.0\% on sets 1 and 2; it also achieves higher joint success at 8B and 32B.
Appendix~\ref{app:access} reports strict scoring and paired differences for all sizes.
CaptionSmiths trained on our COCO split achieves 18.0\% and 17.8\% joint success on the two sets; adding the task instruction gives 11.5\% and 11.6\%.
The 4B SFT+DPO student exceeds the standard CaptionSmiths variant by 42.5\% and 39.5\%, respectively.
CaptionSmiths and our students share the training-image split but use different supervision; Appendix~\ref{app:captionsmiths} gives training details and paired intervals.

\paragraph{Additional prompts and candidate selection.}
On the first image set at 4B, explicitly stating the evaluation rules gives 31.5\% joint success, and adding two training-image demonstrations per condition gives 29.4\%.
The distilled SFT+DPO model outperforms these prompts by 29.0\% and 31.1\%, respectively.
Selecting the first of four prompt-only candidates that passes the same rules gives 59.1\%, compared with 60.5\% after SFT+DPO.
The paired difference is +1.4\% and is uncertain.
Appendix~\ref{app:revision-baselines} gives the selection fallback, strict scoring, and auxiliary comparisons with students trained on 4B-generated descriptions.
On the confirmation set under strict scoring, even selecting the best of four prompt-only candidates with the evaluation rules themselves reaches 45.1\%, below the 48.1\% of a single greedy SFT pass and at higher measured latency (Appendix~\ref{app:oracle}).

%% file: tables/teacher_main.tex
\begin{tabular}{lllccccc}
\toprule
Size & \multicolumn{2}{c}{Method} & Vocabulary & Syntax & Length & Joint (set 1) & Joint (set 2) \\
\midrule
4B & \multirow{3}{*}{Ours} & Prompt & 60.1 & 74.4 & 87.2 & 36.0 \ci{32.9}{39.2} & 35.4 \ci{32.0}{38.8} \\
4B & & SFT & \underline{63.9} & \underline{95.6} & \textbf{96.8} & \underline{59.4} \ci{55.9}{62.9} & \underline{55.6} \ci{52.2}{59.0} \\
4B & & SFT+DPO & \textbf{64.8} & \textbf{96.6} & \underline{95.6} & \textbf{60.5} \ci{57.0}{64.1} & \textbf{57.2} \ci{53.9}{60.6} \\
\noalign{\vskip 1.5pt}
\hdashline[3pt/2pt]
\noalign{\vskip 1.5pt}
4B & \multicolumn{2}{l}{ACCESS \citep{martin-etal-2020-controllable}} & 51.1 & 53.0 & 82.0 & 24.0 \ci{22.5}{25.5} & 23.2 \ci{21.2}{25.2} \\
\midrule
8B & \multirow{3}{*}{Ours} & Prompt & 55.2 & 92.0 & 76.6 & 38.6 \ci{35.9}{41.4} & 38.0 \ci{34.8}{41.4} \\
8B & & SFT & \underline{60.6} & \underline{96.2} & \underline{97.6} & \underline{57.4} \ci{53.9}{61.0} & \underline{53.6} \ci{50.4}{56.9} \\
8B & & SFT+DPO & \textbf{62.8} & \textbf{97.9} & \textbf{98.5} & \textbf{60.5} \ci{56.8}{64.2} & \textbf{60.5} \ci{56.9}{64.0} \\
\noalign{\vskip 1.5pt}
\hdashline[3pt/2pt]
\noalign{\vskip 1.5pt}
8B & \multicolumn{2}{l}{ACCESS \citep{martin-etal-2020-controllable}} & 49.1 & 50.2 & 90.6 & 21.6 \ci{19.9}{23.2} & 22.4 \ci{20.6}{24.1} \\
\midrule
32B & \multirow{3}{*}{Ours} & Prompt & 57.6 & 95.1 & 53.1 & 31.8 \ci{28.1}{35.5} & 32.8 \ci{29.4}{36.1} \\
32B & & SFT & \underline{62.5} & \underline{96.2} & \textbf{97.8} & \underline{59.6} \ci{56.2}{62.9} & \underline{57.6} \ci{54.4}{61.0} \\
32B & & SFT+DPO & \textbf{65.1} & \textbf{97.5} & \underline{97.2} & \textbf{61.6} \ci{58.4}{64.9} & \textbf{59.8} \ci{56.4}{63.2} \\
\noalign{\vskip 1.5pt}
\hdashline[3pt/2pt]
\noalign{\vskip 1.5pt}
32B & \multicolumn{2}{l}{ACCESS \citep{martin-etal-2020-controllable}} & 46.2 & 52.4 & 87.5 & 22.2 \ci{20.1}{24.4} & 21.9 \ci{19.6}{24.2} \\
\noalign{\vskip 1.5pt}
\hdashline[3pt/2pt]
\noalign{\vskip 1.5pt}
7B & \multicolumn{2}{l}{CaptionSmiths\textsuperscript{$\dagger$} \citep{saito-etal-2025-captionsmiths}} & 43.1 & 54.4 & 73.9 & 18.0 \ci{16.4}{19.6} & 17.8 \ci{16.0}{19.5} \\
7B & \multicolumn{2}{l}{\quad + task instruction} & 46.2 & 50.9 & 49.2 & 11.5 \ci{10.5}{12.5} & 11.6 \ci{10.6}{12.8} \\
\noalign{\vskip 1.5pt}
\hdashline[3pt/2pt]
\noalign{\vskip 1.5pt}
125B & \multicolumn{2}{l}{Teacher reference} & 54.1 & 90.1 & 86.9 & 43.2 \ci{40.2}{46.2} & 45.6 \ci{41.9}{49.5} \\
\bottomrule
\end{tabular}

%% file: tables/strict_main.tex
\begin{tabular}{llccc}
\toprule
Size & Method & Set 1 & Set 2 & Confirmation \\
\midrule
4B & Prompt & 22.6 & 20.9 & 24.5 \\
 & SFT & \textbf{48.5} & \underline{45.8} & \underline{48.1} \\
 & SFT+DPO & \textbf{48.5} & \textbf{46.8} & \textbf{48.8} \\
\midrule
8B & Prompt & 24.9 & 25.6 & 24.3 \\
 & SFT & \underline{47.8} & \underline{43.6} & \underline{49.1} \\
 & SFT+DPO & \textbf{51.9} & \textbf{52.0} & \textbf{52.5} \\
\midrule
32B & Prompt & 17.1 & 17.6 & 19.1 \\
 & SFT & \textbf{51.5} & \underline{49.5} & \underline{50.8} \\
 & SFT+DPO & \underline{51.4} & \textbf{52.3} & \textbf{52.5} \\
\bottomrule
\end{tabular}

%% file: 05_cross_axis.tex
\section{Cross-axis Effects in the 4B SFT+DPO Model}
\label{sec:cross-axis}

We examined whether changing one request preserves compliance with the other two in the 4B SFT+DPO student trained on 125B-teacher descriptions (Table~\ref{tab:main}), hereafter the final 4B model.
Using the eight original conditions on each of the two 100-image sets, we changed one request at a time while holding the image and the other requests fixed.
Each operation gives four pairs per image; we averaged their differences within each image, including outputs that fail a constraint.
The following results describe observed changes. Appendix~\ref{app:cross-axis} reports effect sizes, uncertainty, and exploratory tests whose significance depends on the scope of multiple-comparison correction.

\paragraph{Responses to the requested changes.}
Changing basic to advanced vocabulary increases the B1-or-higher share of recognized content words by 19.55\% and 17.54\% on sets 1 and 2, respectively.
Changing simple to complex syntax increases detected clause counts by 1.26 and 1.32; changing short to long increases word counts by 9.94 and 10.09.

\paragraph{Compliance with the fixed requests.}
Changing the length request from short to long reduces compliance with the fixed syntax request by 4.75\% and 5.00\%.
Complex-syntax requests reduce strict length compliance by 11.75\% and 10.50\%; with two-word tolerance, the reductions are 1.75\% and 2.25\%.
Thus, the model moves each intended attribute in the requested direction on average, while some changes reduce compliance with another request in these outputs.

%% file: 07_human_evaluation.tex
\section{Human Evaluation of Difficulty}
\label{sec:human}

We tested whether the final 4B SFT+DPO model's vocabulary and syntax requests correspond to perceived difficulty.
Twenty-four native English speakers recruited through Prolific rated vocabulary difficulty and sentence-structure difficulty separately for an English learner, each from 1 (very easy) to 5 (very difficult).
They evaluated 160 descriptions: all eight original conditions for 20 images from the second evaluation set.
Each participant evaluated one description per image, with three raters per description, giving 480 sentence evaluations.
Model identity, requested conditions, and automatic scores were hidden.
Participants gave consent and were paid at \pounds9 per hour.
Appendix~\ref{app:human} reports demographics, response screening, and the three other questions on clause presence, naturalness, and unsupported content.

We fit linear models with all three requests and their interactions, participant and image fixed effects, and standard errors clustered by participant and image.
``Cannot judge'' responses leave 478 vocabulary and 479 syntax ratings.
Table~\ref{tab:human-primary} reports adjusted differences averaged equally over the other requests, with Holm correction across the two intended contrasts.
The analysis is exploratory: the sample includes three pilot participants, and three other respondents were excluded after response inspection.

\begin{table}[t]
  \centering
  \small
  \setlength{\tabcolsep}{2pt}
  \input{tables/human_primary}
  \caption{Human difficulty ratings (1--5). Vocabulary compares basic $\to$ advanced; syntax compares simple $\to$ complex. Means weight the 160 descriptions equally. Adjusted differences are rating points with pointwise 95\% intervals; $p$ values use Holm correction for two tests. The analysis includes 24 participants and 20 images.}
  \label{tab:human-primary}
\end{table}

Both requests increase their corresponding difficulty ratings.
Advanced vocabulary also increases syntax ratings by 0.27 points, while long requests increase vocabulary and syntax ratings by 0.73 and 0.85 points, respectively (Appendix~\ref{app:human}).
These results support correspondence between the requested adjustments and perceived difficulty, while showing that the two ratings respond to other requests as well.

%% file: tables/human_primary.tex
\begin{tabular}{@{}lccr@{}}
\toprule
Rating & Means & Adj. difference & $p$ \\
\midrule
Vocabulary & 2.15 $\to$ 2.72 & +0.57 [0.32, 0.83] & 0.00034 \\
Syntax & 2.41 $\to$ 2.73 & +0.32 [0.11, 0.53] & 0.00462 \\
\bottomrule
\end{tabular}

%% file: 07b_human_quality.tex
\section{Human Evaluation of Output Quality}
\label{sec:human-quality}

The difficulty study above rates one checkpoint. We therefore ran a second, blinded study comparing the three 4B training stages directly. Sixteen native English speakers recruited through Prolific each rated 40 blinded sentences drawn from the confirmation set (20 images $\times$ eight requests $\times$ three methods: prompting, SFT, and SFT+DPO; 480 distinct sentences, 640 sentence evaluations, about 1.33 raters per sentence). Each sentence received six ratings from 1 to 5: visual faithfulness, coverage of the main information, naturalness, and fit to the requested vocabulary, syntax, and length. Method identity and automatic scores were hidden. Three instruction checks were interleaved; every analysed response passed at least two of the three, and no ratings were missing. Participants gave consent and were paid \pounds2.25 per completed list (\pounds3.00 for the pilot list). Appendix~\ref{app:human} gives the assignment design and screening details.

\begin{table}[t]
  \centering
  \small
  \setlength{\tabcolsep}{3pt}
  \input{tables/human_quality}
  \caption{Mean human ratings (1--5) for the three 4B training stages on confirmation-set outputs. Sixteen raters, 640 sentence evaluations, 3,840 ratings in total.}
  \label{tab:human-quality}
\end{table}

Training changes the ratings little, except where the automatic gains concentrate (Table~\ref{tab:human-quality}). Under the prespecified image-level paired bootstrap, SFT improves length fit over prompting by +0.33 points (95\% CI [+0.18, +0.49]), which shows that the scale registers differences of this size. The remaining SFT--prompting contrasts are small and uncertain: faithfulness +0.08 [$-$0.04, +0.21], main information +0.06 [$-$0.07, +0.19], naturalness +0.01 [$-$0.05, +0.09], vocabulary fit +0.01 [$-$0.16, +0.17], and syntax fit +0.00 [$-$0.09, +0.10]. No SFT+DPO--SFT contrast excludes zero on any of the six ratings. Ratings are concentrated at the top of the scale (76.1\% are 5), so small differences are hard to detect; we read the study as showing that the large automatic compliance gains come with no detectable loss in perceived faithfulness or naturalness, and with a visible gain in length fit. The vocabulary- and syntax-fit ratings also speak to the concern that the students merely satisfy surface rules: raters judge the trained outputs to fit the requested vocabulary and syntax about as well as prompted outputs (4.62--4.63 and 4.70--4.71, respectively).

%% file: tables/human_quality.tex
\begin{tabular}{lccc}
\toprule
Rating (1--5) & Prompt & SFT & SFT+DPO \\
\midrule
Visual faithfulness & 4.54 & 4.63 & 4.65 \\
Main information & 4.68 & 4.74 & 4.79 \\
Naturalness & 4.74 & 4.76 & 4.74 \\
Vocabulary fit & 4.62 & 4.63 & 4.54 \\
Syntax fit & 4.70 & 4.71 & 4.68 \\
Length fit & 4.37 & 4.71 & 4.68 \\
\bottomrule
\end{tabular}

%% file: 07_followup.tex
\section{Additional Evaluations}
\label{sec:followup}

\subsection{Transfer to changed instructions}
\label{sec:transfer}

To test whether the 125B-teacher students follow new word ranges and multi-sentence requests, we evaluate the same checkpoints with changed instructions on 100 additional COCO images.
These images are neither identical nor near duplicates of the 750 images used earlier.
A separate 20-image development pilot fixed the new wording and evaluation rules before final generation.

Each image receives eight combinations of vocabulary (basic or advanced), syntax (simple or complex), and total length (15--20 or 30--40 words).
Short descriptions may contain one to three sentences; long descriptions must contain two to five.
The vocabulary judge estimates a band from distinct recognized content-word types, then compares it with the requested target.
For syntax, simple requires zero finite subordinate or relative clauses, and complex requires at least one. Both exclude absolute constructions (phrases with their own subject, such as ``her arms folded''); simple also excludes sentence-initial participial clauses, and complex excludes subordinate or relative clauses nested inside another such clause.
These rules have not been validated as sentence-level CEFR ratings; Appendix~\ref{app:transfer-rules} gives the prompts and scoring rules.

We score exact length compliance and report \emph{three-axis success} when vocabulary, syntax, and length all pass; success including format is reported separately.
Each method generates 800 outputs using greedy decoding with a 160-token limit; intervals use 10,000 paired image resamples.
Changes to the task, scoring rules, and inference setup preclude direct comparison with the original evaluation (Appendix~\ref{app:transfer-rules}).

\begin{table*}[t]
  \centering
  \small
  \setlength{\tabcolsep}{4pt}
  \input{tables/followup_transfer}
  \caption{Transfer to changed instructions (\%, 800 outputs per row), using the same 125B-teacher student checkpoints as Table~\ref{tab:main}. Length strictly requires 15--20 or 30--40 words. Three-axis success excludes format; the last column adds it. Brackets are 95\% image-bootstrap intervals. Within each size and column, bold marks the best rate and underlining the second best; ties share a rank.}
  \label{tab:transfer}
\end{table*}

SFT and SFT+DPO improve three-axis success over prompting at every size (Table~\ref{tab:transfer}).
At 4B, three-axis success rises from 30.00\% with SFT to 37.00\% with SFT+DPO, a gain of 7.00\%.
At 8B the gain is 5.38\%; the 32B change is $-1.00$\% and is uncertain.
Appendix~\ref{app:transfer-rules} reports the paired differences and confidence intervals.
The 4B gain comes mainly from syntax: from SFT to SFT+DPO, the complex-syntax pass rate rises from 31.50\% to 58.00\%, while advanced-vocabulary compliance falls from 34.75\% to 32.00\%. Each comparison uses the 400 outputs with the corresponding request.

The 32B students have the highest overall vocabulary pass rates, 70.12\% after SFT and 70.50\% after DPO, but only 20.88\% and 16.62\% length compliance.
Exceeding the word ranges limits their three-axis success: 628/800 SFT outputs and 661/800 DPO outputs are too long.
The corresponding DPO counts are 103/800 at 4B and 173/800 at 8B.
At 32B, 33 outputs that passed all three axes after SFT fail after DPO; 27 fail length alone.

\begin{table}[t]
  \centering
  \small
  \setlength{\tabcolsep}{3pt}
  \input{tables/followup_lengths}
  \caption{Word counts after SFT+DPO under changed instructions. Each requested range has 400 outputs per size; cells show median / mean words. The final column is the percentage above the range among all 800 outputs.}
  \label{tab:transfer-length}
\end{table}

The 125B teacher, without task-specific training, also struggles with length: 580/800 outputs exceed the range, and three-axis success is 14.25\%.
Its vocabulary and syntax pass rates are 66.50\% and 77.38\%, respectively.
Differences in architecture and inference software make this an exploratory reference rather than a controlled comparison of student sizes.

Appendix~\ref{app:transfer-examples} presents a source image and example outputs.

\subsection{Sentence-boundary stopping for long outputs}
\label{sec:stopping}

The 32B failure mode above is overshoot: generation continues past sentence ends that already satisfy the request. We therefore test a deterministic decoding-time rule that needs no retraining: stop at the first sentence end at which both the minimum word count and the requested sentence count are met, and otherwise end generation as usual. Applied offline to the saved generations of Table~\ref{tab:transfer}, the rule raises 32B three-axis success by 23.1--30.3 percentage points, with smaller gains at 4B and 8B (Table~\ref{tab:stopping}). The cost is a small drop in vocabulary pass rate (32B SFT+DPO: $-$4.3 points, 95\% CI [$-$6.6, $-$2.0]), because later sentences can carry requested vocabulary; syntax pass rates are nearly unchanged.

\begin{table}[t]
  \centering
  \small
  \setlength{\tabcolsep}{3pt}
  \input{tables/followup_stopping}
  \caption{Three-axis success (\%) under changed instructions, without and with sentence-boundary stopping applied to the same saved generations (100 images, 800 outputs per row).}
  \label{tab:stopping}
\end{table}

Generating with the rule active, on 25 of the images (200 outputs per method), reproduces the offline truncation exactly on the same hardware for all 600 outputs. For 32B SFT+DPO, three-axis success rises from 14.0\% to 42.0\% and median latency falls from 2.60\,s to 2.29\,s. An independent visual audit of the 32B SFT+DPO outputs (30 images, 240 outputs) finds that stopping raises visual claim precision by 1.4 points [+0.2, +2.6] and lowers salient-fact recall by 4.8 points [$-$6.3, $-$3.4], as trailing sentences are removed; at 4B the corresponding changes are smaller (+0.7 [+0.0, +1.6] and $-$0.6 [$-$1.3, +0.2]). Applied to the single-sentence confirmation set, the same rule changes only 2 of 7,200 outputs and no method's strict joint success by more than 0.25 points, so it can remain active for the original task. We treat the rule as a complement to training rather than a substitute: stopped prompting still trails stopped SFT by 12.9 points at 4B and 11.3 points at 8B on the follow-up requests.

\subsection{Describe, then rewrite}
\label{sec:twostage}

We also test two-stage generation with the prompt-only 4B model under the changed instructions.
First, it writes a description of the image without linguistic targets.
It then rewrites that description under each requested condition using only the text.
The eight requests for each image share one first-stage description.
The comparison uses the same 100 images and 800 conditions as direct 4B prompting, with greedy decoding and a 160-token limit for the first stage and each rewrite.
This compares complete generation procedures, including the change from image input to text input when rewriting.

\begin{table}[t]
  \centering
  \small
  \setlength{\tabcolsep}{3pt}
  \input{tables/followup_two_stage}
  \caption{Direct versus two-stage 4B generation under changed instructions (\%, 800 outputs per method). Differences are two-stage minus direct, in percentage points, with 95\% paired image-bootstrap intervals. Within each row, bold marks the best rate and underlining the second best.}
  \label{tab:twostage-transfer}
\end{table}

Three-axis success is 18.25\% with direct prompting and 20.12\% with two-stage generation; the paired gain is uncertain.
Including format, success falls by 6.13\%, with worse word-count and sentence-count compliance (Table~\ref{tab:twostage-transfer}).
The initial descriptions have a median of 76 words; final rewrites average 30.76 words versus 23.45 for direct outputs.
We also measure the fraction of annotated COCO object categories mentioned, pooling category counts across outputs (Appendix~\ref{app:object-mentions}).
This proxy shows no clear advantage: 61.27\% versus 61.11\% on the same 98 images with annotations (784 outputs per method).
It does not measure human-rated information retention or faithfulness.

%% file: tables/followup_transfer.tex
\begin{tabular}{lrrrlr}
\toprule
Model/method & Vocabulary & Syntax & Length & Three-axis success & With format \\
\midrule
4B Prompt & \underline{62.88} & 57.00 & 67.12 & 18.25 \ci{15.75}{20.88} & 16.38 \\
4B SFT & \textbf{63.75} & \underline{63.00} & \textbf{81.38} & \underline{30.00} \ci{27.25}{32.88} & \underline{28.88} \\
4B SFT+DPO & 62.12 & \textbf{76.88} & \underline{81.12} & \textbf{37.00} \ci{34.0}{40.0} & \textbf{35.12} \\
\noalign{\vskip 2pt}
\hdashline[1pt/2pt]
\noalign{\vskip 2pt}
8B Prompt & 61.50 & 60.25 & 56.50 & 20.75 \ci{18.5}{23.12} & 20.25 \\
8B SFT & \textbf{64.88} & \underline{69.25} & \underline{66.25} & \underline{29.12} \ci{26.12}{32.25} & \underline{28.50} \\
8B SFT+DPO & \underline{62.88} & \textbf{78.62} & \textbf{73.38} & \textbf{34.50} \ci{31.37}{37.62} & \textbf{34.25} \\
\noalign{\vskip 2pt}
\hdashline[1pt/2pt]
\noalign{\vskip 2pt}
32B Prompt & 66.12 & \textbf{80.88} & 6.88 & 4.75 \ci{3.38}{6.25} & 3.88 \\
32B SFT & \underline{70.12} & 70.38 & \textbf{20.88} & \textbf{11.50} \ci{9.25}{13.75} & \textbf{11.38} \\
32B SFT+DPO & \textbf{70.50} & \underline{77.25} & \underline{16.62} & \underline{10.50} \ci{8.5}{12.5} & \underline{10.38} \\
\bottomrule
\end{tabular}

%% file: tables/followup_lengths.tex
\begin{tabular}{lrrr}
\toprule
Size & 15--20 words & 30--40 words & Too long (\%) \\
\midrule
4B & 18 / 18.66 & 34 / 33.91 & 12.88 \\
8B & 20 / 20.16 & 34 / 33.90 & 21.62 \\
32B & 25 / 25.46 & 43 / 42.84 & 82.62 \\
\bottomrule
\end{tabular}

%% file: tables/followup_stopping.tex
\begin{tabular}{llcc}
\toprule
Size & Method & No stopping & With stopping \\
\midrule
4B & Prompt & 18.3 & 18.6 \\
 & SFT & 30.0 & 31.5 \\
 & SFT+DPO & 37.0 & 40.0 \\
\midrule
8B & Prompt & 20.8 & 24.1 \\
 & SFT & 29.1 & 35.4 \\
 & SFT+DPO & 34.5 & 38.5 \\
\midrule
32B & Prompt & 4.8 & 35.0 \\
 & SFT & 11.5 & 34.6 \\
 & SFT+DPO & 10.5 & 37.3 \\
\bottomrule
\end{tabular}

%% file: tables/followup_two_stage.tex
\begin{tabular}{lrrr}
\toprule
Criterion & Direct & Two-stage & Difference (pp) \\
\midrule
Three axes & \underline{18.25} & \textbf{20.12} & +1.88 \ci{-1.75}{5.62} \\
With format & \textbf{16.38} & \underline{10.25} & -6.13 \ci{-9.12}{-3.12} \\
Length & \textbf{67.12} & \underline{54.0} & -13.12 \ci{-19.63}{-6.5} \\
Format & \textbf{89.5} & \underline{63.5} & -26.00 \ci{-30.25}{-21.88} \\
\bottomrule
\end{tabular}

%% file: 07_discussion.tex
\section{Discussion}
\label{sec:discussion}

\paragraph{Preserving the other requirements.}
In the final 4B model, long requests produce descriptions with higher shares of recognized B1-or-higher words and more detected clauses (Section~\ref{sec:cross-axis}).
One possible explanation for the additional clauses is that longer descriptions provide more opportunities to include them.
A relative or subordinate clause may itself require additional words.
In the final 4B model, requesting complex syntax adds approximately one word on each set.
The larger loss under strict length scoring suggests that some changes remain compatible with approximate length requirements while making exact adherence harder.

Clause presence and minimum advanced-word counts themselves depend on how many words an output contains.
A change in an automatic indicator can remain within its fixed target, and vocabulary bands and clause counts do not directly measure perceived difficulty.

\paragraph{Interpreting human difficulty ratings.}
Advanced vocabulary requests are associated with higher syntax-difficulty ratings after accounting for the other requests (Section~\ref{sec:human}).
With syntax requests held fixed, the increase could reflect changes in the generated structures, lexical demands affecting syntactic processing, or an impression of difficult vocabulary carrying over to the syntax rating.
Reading-time studies manipulating verb frequency show shifts in when structural processing difficulty appears \citep{tily-etal-2010-timecourse,hopp-2016-timing}.
Distinguishing output changes from processing and judgment effects would require comparisons that hold the generated structure fixed while varying vocabulary, alongside separate measures of processing and ratings.

\paragraph{Length compliance under changed instructions.}
Under changed instructions, advanced vocabulary and complex syntax add 2.69 and 1.78 words, respectively, to long 32B SFT outputs.
Excess length also occurs under easier requests: 190/200 short, simple-syntax outputs exceed 20 words.
Before task-specific training, the 32B model exceeds the requested ranges in 744/800 outputs across all eight conditions.
Supplementary analyses of cross-axis effects, sentence counts, and preference-pair balance appear in Appendix~\ref{app:32b-length}.

\paragraph{Length supervision and future comparisons.}
SFT and DPO provide indirect length supervision rather than an explicit decoding-time word budget.
Among 253 overlong, long-request 32B SFT outputs, 232 (91.7\%) contain a sentence boundary that meets both the requested word range and sentence count, yet generation continues.
Section~\ref{sec:stopping} tests the corresponding budget-aware decoding rule: sentence-boundary stopping recovers much of the 32B loss on changed instructions, at a small cost in vocabulary compliance and salient-fact recall.
Controlled comparisons varying word ranges, sentence-count instructions, and linguistic wording separately would help explain the transfer gap.

%% file: 06_conclusion.tex
\section{Conclusion}

We distill filtered descriptions from a 125B teacher into 4B, 8B, and 32B students that accept separate vocabulary, syntax, and length requests.
Task-specific training improves joint constraint compliance over prompting on both image sets; the additional effect of DPO depends on the model and evaluation rule.
The improvement holds on a frozen confirmation set under strict scoring, and a blinded human study finds no detectable loss in perceived faithfulness or naturalness after training.
Under changed instructions, a deterministic sentence-boundary stopping rule substantially reduces the 32B model's overshoot without retraining.
The final 4B model responds to changes in each requested attribute, but changing one request can reduce compliance with another, particularly under strict length scoring.
Under changed instructions, the 32B model often exceeds the new word ranges despite high vocabulary and syntax pass rates.
For the prompt-only 4B model, describing before rewriting gives no clear three-axis advantage over direct prompting.
Human ratings of the final 4B model increase with the corresponding vocabulary and syntax requests, while also reflecting cross-axis effects.

%% file: 08_limitations.tex
\section*{Limitations}

Our targets are operational definitions of vocabulary, syntax, and length.
We evaluate generation under supplied targets; selecting targets for individual readers remains outside the evaluation.
Word-list membership depends on lexical coverage and part-of-speech analysis; it does not capture word senses, multiword expressions, or sentence-level CEFR proficiency.
The advanced-vocabulary prompt asks for \emph{a few} more advanced words, whereas the evaluator requires at least 35\% of recognized content words at B1 or above.
The clause detector does not measure all aspects of syntactic difficulty.
Appendix~\ref{app:additional} shows how conclusions change under exact adherence to the requested ranges.

In the original evaluation, the same automatic rules select training targets, label preferences, select SFT checkpoints, and score outputs.
Automatic pass rates measure generalization to new images under those rules.
The human evaluation independently assesses perceived difficulty and collects naturalness and unsupported-content judgments for the final 4B SFT+DPO checkpoint on 20 images.
Its 24 native English speakers judge difficulty for learners; learner comprehension and learning outcomes remain unmeasured.
The sample includes pilot responses and post-hoc exclusions, and inference uses a linear approximation to ordinal ratings with only 24 participant and 20 image clusters.
The blinded quality study (Section~\ref{sec:human-quality}) compares the three 4B training stages, but with about 1.33 raters per sentence and ratings concentrated at the top of the scale (76.1\% are 5), it can miss small quality differences; no human comparison across model sizes was conducted.
The training-data faithfulness screen is model-based and may retain unsupported details; Appendix~\ref{app:screen-audit} quantifies its errors with an independent judge.
The visual audit of Appendix~\ref{app:visual-audit} evaluates the 125B-teacher students, but it relies on a single automated judge; its labels are not calibrated human error probabilities, and disagreement between judges does not establish which judgment is correct.
A separate quality screen covers 4B outputs on 30 images from prompt-only baselines and students trained on 4B-generated descriptions, and flags more unsupported content after SFT+DPO.
Captioning research shows that better standard sentence-level scores need not imply fewer hallucinated objects \citep{rohrbach-etal-2018-object}.
Our linguistic pass rates likewise require separate assessment of visual correctness, content preservation, and educational benefit.

The study uses English descriptions, COCO photographs, and one VLM family for the student models.
COCO images may have appeared in pretraining.
Both original evaluation sets informed exploratory work.
The confirmation set was frozen before generation and provides an untouched evaluation for our pipeline, but its images come from the same COCO pool and may have appeared in pretraining.
Most conditions have one training seed, and bootstrap intervals quantify image variation only.
The repeated-run check in Appendix~\ref{app:4b-source} covers only 4B SFT trained on 4B-generated descriptions; variation in the 125B-teacher students, DPO, and larger models remains unmeasured.
Appendix~\ref{app:diversity} describes how variable the confirmation-set outputs themselves are.
Automatic-evaluation intervals and the human-rating intervals are pointwise; the human-rating tests use Holm correction within the specified families.

The comparison between 125B- and 4B-generated training descriptions also changes data volume, inference software, and the source of DPO candidates.
It cannot isolate teacher capacity.
In the auxiliary 4B-generated-data experiment, both DPO variants reuse 4B-generated pairs at larger sizes, and changing pair selection alters vocabulary and length directions together.
Grammar directions remain unequal in both training experiments.
We have not isolated these factors, compared length-regularized preference objectives, or established an optimal training procedure.
The transfer evaluation changes instructions, target rules, word ranges, permitted sentence counts, images, and inference environment together, preventing attribution to any single change (Section~\ref{sec:transfer}).
The original and changed-instruction outcomes use different length rules and different treatment of format, and should not be pooled.

On-device use motivates the study; we measure per-output latency on one Apple Silicon device (Appendix~\ref{app:oracle}), but not device memory or energy use.
Training cost is reported in Appendix~\ref{app:training}.

\section*{Ethical Considerations}

This study uses existing COCO images, automatically generated descriptions, and human ratings.
Both human evaluations received institutional ethics approval (identifying details withheld for anonymous review).
Participants gave consent; difficulty-study participants were paid at \pounds9 per hour through Prolific, and quality-study participants \pounds2.25 per completed list (\pounds3.00 for the pilot list).
We report aggregate demographics and results without participant identifiers.
Descriptions can contain incorrect details, stereotypes, or inappropriate language despite satisfying all measured constraints.
Materials intended for learners should therefore receive human review of content and suitability.
Our rule-based vocabulary labels should not be used to assess a learner's ability or to make high-stakes educational decisions.

We will release the training and evaluation code, the LoRA adapters, image-split identifiers, and evaluation outputs.
Filtered teacher captions and auxiliary artifacts will be released as far as the base-model, teacher, COCO, and lexicon licences permit; raw human ratings are withheld to protect participants.

\paragraph{AI assistance.}
We used OpenAI Codex primarily to support data analysis, including analysis code and interpretation of results.
These tools also assisted with initial manuscript preparation.
The authors take responsibility for the analyses, interpretations, and final manuscript.

%% file: 09_appendix.tex
\section{Prompts and Evaluation Rules}
\label{app:rules}

\subsection{Generation prompt}

The model receives one image and the following user message through its native chat template. Braced fields are replaced with the corresponding instructions below; blank lines follow the original prompt.

\begin{quote}\small
Describe the main visible scene in this image in one English sentence.

Vocabulary: \{vocabulary\}\newline
Grammar: \{grammar\}\newline
Length: Write \{length\} words.

Keep the description natural and faithful to the image.\newline
Return only the sentence.
\end{quote}

\paragraph{Vocabulary instructions.}
Basic: ``Use basic, common everyday words.'' Advanced: ``Use a few more advanced, precise content words where natural, without changing the meaning.''

\paragraph{Grammar instructions.}
Simple: ``Use a simple sentence without relative or subordinate clauses.'' Complex: ``Include a relative clause (with who, which, or that) or a subordinate clause (with a word such as because, while, when, or as) that has its own subject and verb.''

\paragraph{Length instructions.}
The length field is ``10--14'' or ``20--24'', with an en dash in the actual prompt. All eight combinations specify all three axes; there are no conditions with an unspecified axis.

\subsection{Vocabulary}

The evaluator uses spaCy 3.8.16 \citep{honnibal-etal-2020-spacy} with Explosion's \texttt{en\_core\_web\_trf} 3.8.0 \citep{explosion-en-core-web-trf}. The pipeline uses RoBERTa-base \citep{liu-etal-2019-roberta}; its model metadata credits OntoNotes 5.0 \citep{weischedel-etal-2013-ontonotes}, Emory University's ClearNLP dependency conversion \citep{emory-clearnlp-conversion}, and Princeton WordNet 3.0 \citep{miller-1995-wordnet}.
The evaluator considers alphabetic tokens tagged as nouns, verbs, adjectives, or adverbs; proper nouns, auxiliaries, and numbers are excluded. We combine CEFR-J Wordlist 1.6 (A1--B2; \citealp{tono-2020-cefrj}) and Octanove C1/C2 1.0 \citep{octanove-vocabulary}. Where entries overlap, the lowest listed level is retained for each word--part-of-speech pair. Lookup first tries the lemma with its part of speech, then the surface form with that part of speech, and finally the surface form under any part of speech. It does not disambiguate word senses or recognize multiword expressions.

Let $C(y)$ be the number of eligible content-word occurrences, $K(y)$ the number recognized by the lexicons, and $A(y)$ the number of recognized occurrences at B1 or above. When $K(y)>0$, the advanced-word ratio is
\begin{equation}
 r(y)=\frac{A(y)}{K(y)}.
\end{equation}
A basic target passes if $r(y)\leq0.25$. An advanced target passes if $r(y)\geq0.35$ and $A(y)\geq2$. Both require $C(y)>0$, $K(y)>0$, and $K(y)/C(y)\geq0.5$. An output with insufficient coverage is undetermined and fails joint success; other outputs meeting neither threshold fail both vocabulary targets. Thus the labels are not complementary, and unknown words are excluded from the ratio rather than automatically counted as advanced. Token occurrences, including repeated occurrences, enter the numerator and denominator.

These rules operationalize relative lexical demands. The advanced instruction's ``a few'' does not explicitly communicate the 35\% threshold, and neither threshold establishes a sentence-level CEFR classification. A learner study would be required to establish whether the labels predict comprehension or learning.

\subsection{Grammar, length, and format}

The grammar detector examines dependency labels \texttt{relcl}, \texttt{advcl}, \texttt{ccomp}, \texttt{csubj}, \texttt{csubjpass}, and \texttt{acl}. A candidate clause counts if its head, or an auxiliary child, has the morphological feature \texttt{VerbForm=Fin}. A relative clause also counts if its subtree contains a preceding relativizer: \emph{who, whom, whose, which, that, where, when}, or \emph{why}. This latter heuristic need not independently confirm a finite verb. Simple requires zero counted clauses; complex requires at least one. Coordination, open clausal complements, and nonfinite phrases without the qualifying relative-clause pattern do not count. Parser errors and the absence of an upper bound on clause count limit interpretation as grammatical difficulty.

For length, text is split at whitespace, en/em dashes, and slashes. A resulting token counts if it contains an ASCII letter or digit. Contractions and hyphenated compounds count as one word; punctuation-only tokens do not count. This is a word count, not a model-token count. The requested intervals are inclusive: $[10,14]$ and $[20,24]$. The primary evaluation uses the requested intervals without tolerance. A secondary variant allows a two-word tolerance, accepting $[8,16]$ and $[18,26]$. Training targets and checkpoint selection use the strict intervals; neither scoring variant changes the prompt or other evaluation rules.

The format check rejects empty outputs, internal newlines, initial bullets or numbered-list markers, common meta-text prefixes (such as ``Here is'' or ``Answer:''), missing final sentence punctuation, and outputs that did not reach an end-of-sequence token before the generation limit. A punctuation-based heuristic detects multiple sentences, with exemptions for initials and several titles. The language check requires at least one Latin letter and rejects kana and CJK ideographs; it is not a comprehensive English-language detector.

Writing the four pass indicators as $F$, $V$, $G$, and $L$, joint success for image $i$, condition $c$, and output $y_{ic}$ is
\begin{equation}
 J_{ic}=F(y_{ic})V(y_{ic},c)G(y_{ic},c)L(y_{ic},c).
\end{equation}
For $n$ images and all eight conditions, the reported rate is
\begin{equation}
 \widehat{J}=\frac{1}{8n}\sum_{i=1}^{n}\sum_{c=1}^{8}J_{ic}.
\end{equation}
Axis-specific rates use the same $8n$ outputs. Undetermined vocabulary decisions are failures, not missing observations. Neither the automatic faithfulness score nor human quality judgments enter $J_{ic}$.

\section{Data and Training Details}
\label{app:training}

\subsection{Image splits, decoding, and computing resources}

COCO image identifiers are sorted, shuffled with seed 42, and assigned in order to training (500), development (50), and the first evaluation split (100). Exact file duplicates and near duplicates within a Hamming distance of four under a 64-bit average image hash are excluded during selection. The development pilot is the first 20 development images. A second 100-image evaluation split uses seed 2027 and excludes previous identifiers and images meeting the same duplicate criterion. This filtering cannot exclude all related photographs or pretraining contamination. Held-out status refers only to our task-specific training images. Both evaluation splits have been inspected, so the second split does not constitute an untouched confirmatory test.

Images are converted to RGB and downscaled, preserving aspect ratio, to at most 262,144 pixels before model processing. Students and their unadapted prompt controls use bfloat16 through Transformers. Candidate sampling uses temperature 0.7 and nucleus probability 0.9; the local student sampler disables top-$k$ filtering. Four candidates are generated initially, followed by four more for prompts with no passing candidate. Evaluation uses one greedy completion per prompt, with a maximum of 96 new tokens. The processor uses left padding for batched inference. Each complete evaluation set contributes 800 outputs per model and method.

\paragraph{Student models and execution.}
Student loading and inference use Transformers \citep{wolf-etal-2020-transformers}, with Accelerate for automatic placement across GPUs \citep{gugger-etal-2022-accelerate}.
The configured Qwen3-VL-Instruct pretrained revisions are \nolinkurl{ebb281ec70b05090aa6165b016eac8ec08e71b17} (4B), \nolinkurl{0c351dd01ed87e9c1b53cbc748cba10e6187ff3b} (8B), and \nolinkurl{0cfaf48183f594c314753d30a4c4974bc75f3ccb} (32B). Early 4B prompt records lack model and revision fields, so those records alone cannot verify the historical weight identity. Prompt controls and trained students share image preprocessing, the native chat template, prompts, and decoding settings. The archived prompt controls ran on Mac with the MPS backend; the main trained students ran on NVIDIA A100 GPUs through CUDA. Evaluation keeps the trained adapters unmerged. The requested generation settings are aligned, but the numerical backend is not controlled across these runs.

\paragraph{Teacher model and execution.}
The teacher is Qwen Team's Qwen3.8-Flash-Next \citep{qwen2026design}, using Q5\_K\_M GGUF weights distributed by bartowski \citep{bartowski-qwen38-gguf}, recorded weight revision \texttt{928589fd}, and llama.cpp CUDA revision \texttt{448147d} \citep{ggml-llamacpp}.
The A100 teacher-generation script loads the four-part weight set starting at \nolinkurl{Qwen3.8-Flash-Next-Q5_K_M-00001-of-00004.gguf} and the image projector \nolinkurl{mmproj-Qwen3.8-Flash-Next-f16.gguf}.
We follow the original model card's 125B designation (6B active), which excludes its additional 51B n-gram embeddings and 4B multi-token-prediction module.\footnote{\url{https://huggingface.co/Qwen/Qwen3.8-Flash-Next}}
The server loads all model layers onto two GPUs, uses an f16 image projector, disables reasoning, and has a 16,384-token context setting with eight parallel slots. Candidate generation and the direct teacher reference use this same deployment. Teacher sampling requests temperature 0.7 and top-$p$ 0.9; direct evaluation requests temperature zero and top-$p$ 1.0, with a 96-token maximum in both modes. The API client does not explicitly set top-$k$. Images undergo an additional JPEG encoding at quality 92 for API transmission. These differences in model family, quantization, backend, and image transmission mean that teacher--student comparisons do not isolate parameter count.

\paragraph{Hardware and computation time.}
The main CUDA training and evaluation pipeline used NVIDIA A100-SXM4 GPUs with 80\,GB of memory per GPU.
The 4B and 8B students used one GPU per run; the 32B student and 125B teacher used two.
The logged main and additional experiments consumed approximately 177 A100 GPU-hours (Table~\ref{tab:compute-budget}).
For the 4B student, post-training after the shared teacher generation took 13.50 A100 GPU-hours: 5.51 for SFT, 6.86 for student candidate generation and screening, and 1.14 for DPO.
We estimate each stage's cost as its elapsed wall-clock time multiplied by its configured GPU count, covering generation, training, screening, and evaluation, including CPU processing and loading within timed stages.
The additional budget covers teacher generation and 4B retraining under changed instructions, a separate trial from the evaluation of existing checkpoints in Section~\ref{sec:followup}.
Separately, recorded runs used 51.2 device-hours on the Apple MPS backend with an Apple M5 Max and 128\,GB unified memory, and 1.05 GPU-hours for reference inference on one NVIDIA RTX PRO 6000 Blackwell Max-Q with 96\,GB memory.
The RTX reference runs used the Strata inference engine \citep{strata2026} (0.1.40, revision \texttt{82f46a8}) to serve Qwen3.8-Flash-Next with IQ3\_S weights for greedy and best-of-four evaluations.
These figures cover recorded runs; unlogged exploratory trials are excluded.

\begin{table}[t]
\centering
\small
\setlength{\tabcolsep}{3pt}
\begin{tabular}{p{0.64\columnwidth}rr}
\toprule
Logged stage & GPUs & GPU-h \\
\midrule
Main: teacher generation and evaluation & 2 & 37.44 \\
Main: screening and target preparation & 1 & 0.50 \\
Main: 4B training and evaluation & 1 & 14.40 \\
Main: 8B training and evaluation & 1 & 18.81 \\
Main: 32B training and evaluation & 2 & 27.80 \\
\midrule
Main subtotal & & 98.95 \\
Additional teacher generation and 4B retraining under changed instructions & 1--2 & 77.80 \\
\midrule
Total & & 176.76 \\
\bottomrule
\end{tabular}
\caption{Recorded A100 computation budget. Student stages include candidate generation, screening, training, and evaluation. Additional stages use two GPUs for the teacher and one for the student. The total is calculated before rounding the components. Apple MPS and RTX times are reported separately in the text.}
\label{tab:compute-budget}
\end{table}

\subsection{Automatic target selection}

For the main experiment, SFT candidates come from the 125B teacher. In addition to passing all evaluation checks with strict length ranges, an eligible candidate must pass a repetition filter and score at least 0.5 on the following image-conditioned question, asked of the unadapted Qwen3-VL-4B-Instruct model:

\begin{quote}\small
Here is a sentence about this image:\newline
``\{sentence\}''

Does the sentence describe only things that are clearly visible in the image, without adding objects, feelings, intentions, or causes that are not shown? Answer Yes or No.
\end{quote}

The score is $p_{\mathrm{yes}}=P(\mathrm{Yes})/[P(\mathrm{Yes})+P(\mathrm{No})]$ from the first answer-token distribution, summing the token forms with and without a preceding space. It is an uncalibrated screening proxy, not an estimate of human agreement. The repetition filter removes a small fixed list of function words, then flags any remaining word occurring at least three times or any bigram occurring at least twice. Among eligible candidates for each image and condition, we select the one with highest $p_{\mathrm{yes}}$. Seeded subsampling caps every condition at the smallest available count, producing 219 examples per condition and 1,752 in total without duplicating examples. All students use this same sentence-target dataset. The SFT loss uses the retained token sequences; it does not require teacher token probabilities.

DPO candidates come from each student's own selected SFT checkpoint on the 4,000 training image--condition prompts. The chosen response passes every strict rule and the automatic screen. A rejected response must be well formed, pass the same screen, and violate exactly one control axis; an undetermined vocabulary result is not an admissible vocabulary violation. Vocabulary and grammar rejections must differ from the chosen response by no more than three words. Every pair requires a Jaccard overlap of at least 0.2 between content-lemma sets. The admissible rejection with highest overlap is selected, with at most one pair per prompt and violated axis. Preference labels come from these rules and screening procedures, not from the 125B teacher.

We cap every violated axis at the smallest available axis count and divide each axis equally between directions where candidate availability permits. Table~\ref{tab:main-pair-counts} lists the resulting datasets. Vocabulary direction denotes the requested target: a rejected response may lie between the basic and advanced thresholds rather than at the opposite level. Grammar directions remain unequal because there are too few candidates in one direction. The DPO examples and their counts differ across student sizes, so comparisons between sizes include these data differences.

\begin{table}[t]
\centering
\small
\begin{tabular}{lrrr}
\toprule
Pair count & 4B & 8B & 32B \\
\midrule
Total & 576 & 462 & 426 \\
Per violated axis & 192 & 154 & 142 \\
Vocabulary: basic & 96 & 77 & 71 \\
Vocabulary: advanced & 96 & 77 & 71 \\
Length: too short & 96 & 77 & 71 \\
Length: too long & 96 & 77 & 71 \\
Grammar: simple & 107 & 80 & 81 \\
Grammar: complex & 85 & 74 & 61 \\
\bottomrule
\end{tabular}
\caption{Main-experiment DPO training pairs. Vocabulary and grammar rows identify the requested target; length rows identify the rejected response's violation. Each student generates its own candidates after SFT on the same 1,752 teacher descriptions.}
\label{tab:main-pair-counts}
\end{table}

\subsection{Optimization}

SFT and DPO use custom PyTorch training loops \citep{paszke-etal-2019-pytorch} with PEFT adapters \citep{mangrulkar-etal-2022-peft}.
LoRA adapters have rank 16, scale 32, and SFT dropout 0.05. They modify the language model's attention projections ($q,k,v,o$) and feed-forward projections (gate, up, down); the vision encoder and vision--language merger remain frozen. Only assistant answer tokens and their end-of-turn token receive training loss. Image, user-prompt, and other template tokens are masked.

SFT minimizes answer-token cross-entropy with AdamW, learning rate $10^{-4}$, zero weight decay, cosine decay, and a 3\% warmup fraction. The microbatch size is one, with 16-step gradient accumulation and gradient clipping at norm 1. The configured three epochs on 1,752 examples give 327 optimizer steps; the implementation omits the final eight examples in each epoch after shuffling. Each epoch is evaluated greedily on all 400 development outputs, and the checkpoint with highest strict joint success is retained; ties favor the later epoch. These settings are shared across student sizes.

DPO initializes the policy from that SFT checkpoint and uses a frozen copy as the reference $\pi_{\mathrm{ref}}$. For a shared image and prompt $x$, define
\begin{equation}
 \delta_{\theta}(x,y)=\log\frac{\pi_{\theta}(y\mid x)}{\pi_{\mathrm{ref}}(y\mid x)}.
\end{equation}
For chosen response $y^+$ and rejected response $y^-$, the loss is
\begin{equation}
 -\log\sigma\!\left(\beta\left[
 \delta_{\theta}(x,y^+)-\delta_{\theta}(x,y^-)
 \right]\right).
\end{equation}
Sequence log probabilities sum over answer tokens, including the end-of-turn token, without length normalization. We use $\beta=0.1$, AdamW with learning rate $10^{-5}$ and zero weight decay, cosine decay, a 10\% warmup fraction, and the same effective batch size and gradient clipping. Dropout is disabled during DPO. The configured one epoch gives 36, 29, and 27 optimizer steps for 4B, 8B, and 32B, respectively; the final incomplete accumulation group is normalized by its actual size. The final adapter is evaluated without checkpoint selection. All main SFT and DPO runs use seed 42, with one run per condition.

\subsection{Supplementary comparison with 4B-generated training data}
\label{app:small-teacher-method}

The supplementary training-data comparison uses the unadapted 4B model to generate SFT targets under the same prompts and selection rules. It retains 1,384 sentences, with 173 per condition, shared by all three student sizes. Its selected 4B SFT model then generates 22,768 candidates across the 4,000 training prompts. The admissible pool contains 633 vocabulary, 224 grammar, and 783 length pairs. The resulting SFT+DPO condition uses 672 pairs, with 224 per axis: 112 basic and 112 advanced vocabulary targets, 112 too-short and 112 too-long length rejections, and 201 complex and 23 simple grammar targets. All student sizes reuse these 4B-generated pairs.

\emph{DPO with original pairs} uses an alternative set of 672 pairs from the same pool, with 224 per axis but without requiring equal vocabulary and length direction counts. Its length pairs reject 164 too-short and 60 too-long responses; the grammar counts remain 201 complex and 23 simple. The two pair conditions have the same selection criteria and optimization settings. SFT uses 258 optimizer steps across three epochs and selects epoch three at each size; DPO uses 42 steps. Additional SFT runs with seeds 43 and 44 apply only to the 4B-generated-data experiment. These repeated runs do not establish seed stability for the main 125B-teacher experiment.

Relative to this supplementary experiment, the main experiment changes the SFT generator and dataset size, generates preference candidates separately for each student, and uses a different hardware backend.

\subsection{Bootstrap details}

For the original-instruction evaluations, the bootstrap uses 10,000 image resamples with seed 20261008 and percentile endpoints at 2.5\% and 97.5\%; paired comparisons use the same sampled images for both methods. Limitations discusses the shared evaluation rules and sources of uncertainty beyond image sampling.

%% file: 10_additional_results.tex
\section{Additional Results}
\label{app:additional}

In the result tables, bold marks the best displayed rate and underlining the
second-best distinct rate; ties at the displayed precision share a rank.
Method comparisons and length-sensitivity tables rank methods within each
model size. Condition-level joint-success tables compare methods within each
size and condition; axis-level tables compare model sizes for the same axis
and condition. Prompt-baseline tables rank methods within each success column.
For the quality flags, lower rates rank higher. Emphasis denotes numerical
rank, not statistical significance.

\paragraph{Paired training differences.}
\label{app:training-differences}
With two-word tolerance, the 4B SFT+DPO gains over prompting are 24.5 pp [$+20.4$, $+28.6$] on set 1 and 21.9 pp [$+17.8$, $+26.2$] on set 2.
At 8B, the SFT+DPO gains over SFT are 3.1 pp [$+0.4$, $+5.9$] and 6.9 pp [$+4.4$, $+9.4$], respectively.
Brackets give 95\% paired image-bootstrap intervals; each comparison uses 100 images and 800 outputs per method.

\subsection{Strict length sensitivity}

Table~\ref{tab:sensitivity} compares exact adherence to the requested ranges (strict joint success, our primary metric) with the two-word-tolerance variant used in Table~\ref{tab:main}.
Both score the same saved outputs; vocabulary, syntax, format, and denominators are unchanged.
The distilled students improve over prompting at all three sizes on both sets under either rule.
Under strict scoring, DPO improves 8B joint success by 4.13 pp [$+1.25$, $+6.88$] on the first set; the changes at 4B and 32B are uncertain.
On the confirmation set, the paired strict gains of SFT over prompting are 23.6 pp [19.6, 27.6] at 4B, 24.9 pp [20.8, 29.1] at 8B, and 31.6 pp [28.1, 35.1] at 32B; the further gains of SFT+DPO over SFT are 0.6 pp [$-$2.8, 4.0], 3.4 pp [0.1, 6.8], and 1.8 pp [$-$1.0, 4.6], respectively.

\begin{table*}[t]
  \centering
  \small
  \setlength{\tabcolsep}{3pt}
  \input{tables/teacher_sensitivity}
  \caption{Joint success (\%) for the 125B-teacher students under strict length ranges (10--14 or 20--24 words) and two-word tolerance (8--16 or 18--26). Each cell uses the same 800 outputs per model and image set. Brackets are 95\% image-bootstrap intervals.}
  \label{tab:sensitivity}
\end{table*}

\subsection{Condition-level results}

Tables~\ref{tab:conditions1} and \ref{tab:conditions2} retain each vocabulary--syntax--length combination for the distilled students.
Each cell uses 100 outputs, one per image. Success for one combination does not imply reliable control for all combinations.
Tables~\ref{tab:axis-conditions1} and \ref{tab:axis-conditions2} separate the three axis pass rates after SFT+DPO.
Across model sizes and conditions, joint success ranges from 32--84\% on the first set and 32--83\% on the second; vocabulary compliance ranges from 33--91\% and 36--86\%, respectively.
Control therefore remains uneven across requests despite higher overall compliance.

\begin{table*}[t]
  \centering
  \small
  \setlength{\tabcolsep}{3pt}
  \input{tables/teacher_conditions}
  \caption{Joint success (\%) by condition for the 125B-teacher students on the first image set, with two-word length tolerance. P: prompt only; S: SFT; D: SFT+DPO. Each cell contains 100 outputs.}
  \label{tab:conditions1}
\end{table*}

\begin{table*}[t]
  \centering
  \small
  \setlength{\tabcolsep}{3pt}
  \input{tables/teacher_conditions_test2}
  \caption{The same distilled-student comparison on the second image set (100 outputs per cell), with abbreviations and tolerance as in Table~\ref{tab:conditions1}.}
  \label{tab:conditions2}
\end{table*}

\begin{table*}[t]
  \centering
  \small
  \setlength{\tabcolsep}{3pt}
  \input{tables/teacher_axis_conditions}
  \caption{Axis pass rates (\%) for the 125B-teacher students after SFT+DPO, by requested condition on the first image set. V: vocabulary; G: grammar (syntax); L: length with two-word tolerance. Each model contributes 800 outputs across eight conditions, with 100 outputs per cell.}
  \label{tab:axis-conditions1}
\end{table*}

\begin{table*}[t]
  \centering
  \small
  \setlength{\tabcolsep}{3pt}
  \input{tables/teacher_axis_conditions_test2}
  \caption{Axis pass rates (\%) for the 125B-teacher students after SFT+DPO on the second image set. V: vocabulary; G: grammar (syntax); L: length with two-word tolerance. Each model contributes 800 outputs across eight conditions, with 100 outputs per cell.}
  \label{tab:axis-conditions2}
\end{table*}

\subsection{Comparison with 4B-generated training data}
\label{sec:teacher}

An earlier training experiment uses 1,384 filtered descriptions from the prompt-only 4B model (173 per condition) and shares 4B-generated preference pairs across student sizes.
With these targets, the 4B SFT student reaches 44.9\% joint success on the first set and 43.2\% on the second, compared with 59.4\% and 55.6\% using the 125B teacher's targets.
The larger students also obtain higher SFT success with the 125B targets; the strict-rule differences are positive at every size on both sets (Tables~\ref{tab:sensitivity} and \ref{tab:smallteacher-sensitivity}).
This comparison evaluates complete training procedures; Appendix~\ref{app:small-teacher-method} details the differences in data and execution.
Appendix~\ref{app:4b-source} retains the full 4B-generated-data results, including the alternative preference-pair selection.

\subsection{Training on 4B-generated descriptions}
\label{app:4b-source}

\begin{table*}[t]
  \centering
  \small
  \input{tables/short_main}
  \caption{Constraint pass rates (\%) for the auxiliary experiment using 4B-generated descriptions. Vocabulary, syntax, and length columns use the first image set; joint success includes all three and format, with results for both sets. Length allows two words outside the requested range. Each cell uses 800 outputs from 100 images and eight conditions. Brackets are 95\% image-bootstrap intervals; differences use unrounded rates. DPO always follows SFT; ``orig.'' denotes the original pairs, which differ in length and vocabulary composition.}
  \label{tab:smallteacher}
\end{table*}

\paragraph{Training improves pooled joint success.}
SFT improves joint success over prompting at every model size on both image sets (Table~\ref{tab:smallteacher}).
On the first set, the gains are 8.9, 11.9, and 22.8 percentage points (pp) for 4B, 8B, and 32B.
SFT+DPO raises success further, to 53.8\%, 61.1\%, and 61.5\%.
Relative to prompting, these are gains of 17.8, 22.5, and 29.8 pp.
The corresponding rates on the second set are 52.9\%, 55.0\%, and 59.2\%, again exceeding prompting and SFT at every size.
Image-paired 95\% intervals for all six SFT+DPO versus SFT comparisons exclude zero.
These pooled gains hold across sizes and sets, but not every requested combination improves (Tables~\ref{tab:smallteacher-conditions1} and \ref{tab:smallteacher-conditions2}).

\paragraph{Variation across SFT runs.}
Three 4B SFT runs trained on 4B-generated descriptions achieve 30.1\%--32.0\% joint success under strict length scoring on the first image set (100 images, 800 outputs per run).
The runs use seeds 42, 43, and 44 with the training settings in Appendix~\ref{app:small-teacher-method}.

\paragraph{Better syntax does not ensure a joint gain.}
DPO with original pairs improves syntax compliance over SFT at every size, but the joint gains differ.
At 4B, joint success increases from 44.9\% to 51.0\% on the first set.
At 32B, syntax compliance rises from 90.1\% to 95.6\%, while length compliance falls from 96.0\% to 86.2\%.
Joint success is consequently similar: 54.5\% for SFT and 55.0\% for DPO with original pairs, a paired difference of +0.5 pp [95\% CI: $-3.5$, $+4.5$].
The second set likewise gives a small, uncertain joint difference (+1.1 pp [$-2.5$, $+4.8$]).
This does not establish equivalence, but it provides no clear evidence of a joint gain at 32B.
Table~\ref{tab:smallteacher-sensitivity} gives the corresponding results under strict length scoring.

\begin{figure}[t]
  \centering
  \includegraphics[width=\linewidth]{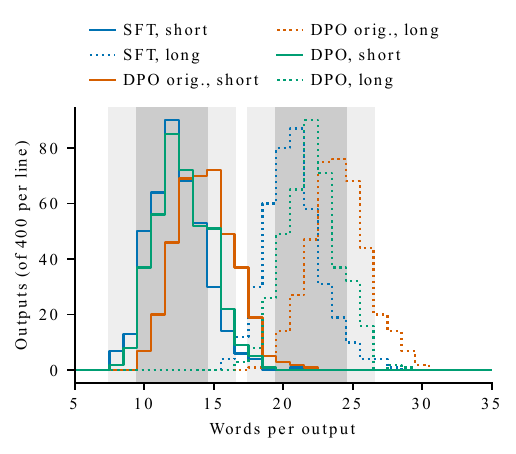}
  \caption{Word counts for 32B outputs on the first set. DPO denotes SFT+DPO trained on 4B-generated descriptions; DPO orig. uses the original pairs. Solid and dotted lines denote short and long requests (400 outputs each). Dark bands show requested ranges; light bands include the two-word tolerance. DPO with original pairs shifts outputs toward longer sentences.}
  \label{fig:length32b}
\end{figure}

\paragraph{Preference directions help explain the trade-off.}
The original pairs contain more too-short than too-long rejections (164 versus 60), so length comparisons more often favor a longer response.
Figure~\ref{fig:length32b} shows the associated change in generated lengths.
The SFT+DPO method in this auxiliary experiment uses equal counts in each length and vocabulary direction, reaching 97.4\% length compliance and 61.5\% joint success at 32B.
Relative to DPO using the original pairs, the joint gain is +6.5 pp [$+3.3$, $+9.8$] on the first set and +8.0 pp [$+4.9$, $+11.0$] on the second.
The 8B model also improves on both sets.
At 4B, the additional gain is smaller and is not clear on the second set (+0.1 pp [$-2.6$, $+2.9$]).
Because vocabulary directions also change and each DPO condition has one training run, this comparison does not isolate the effect of using equal numbers of too-short and too-long rejections.

\subsection{Strict length sensitivity with 4B-generated targets}

Table~\ref{tab:smallteacher-sensitivity} applies the strict-versus-tolerance comparison of Table~\ref{tab:sensitivity} to the 4B-generated-data students. The comparison between prompting and SFT+DPO remains positive at all three sizes on both sets under either rule.

Within this 4B-generated-data experiment, the comparison between DPO with original pairs and SFT is more sensitive. At 32B on the first set, strict joint success falls from 43.0\% to 36.5\%, whereas evaluation with the two-word tolerance gives 54.5\% and 55.0\%. Similar success under tolerance can therefore coexist with more failures under exact requested ranges. SFT+DPO reaches 49.4\% under the strict rule and 61.5\% with the two-word tolerance.

\begin{table*}[t]
  \centering
  \small
  \setlength{\tabcolsep}{3pt}
  \input{tables/short_sensitivity}
  \caption{Joint success (\%) for students trained on 4B-generated descriptions, under strict length ranges (10--14 or 20--24 words) and two-word tolerance (8--16 or 18--26). Each cell uses the same 800 outputs per model and image set. Brackets are 95\% image-bootstrap intervals. ``Orig.'' denotes DPO with original pairs.}
  \label{tab:smallteacher-sensitivity}
\end{table*}

\subsection{ACCESS rewriting baseline}
\label{app:access}

For each source size (4B, 8B, or 32B), the unadapted Qwen3-VL-Instruct model generates one caption per image using greedy decoding and a 96-token limit.
That caption supplies the text input for all eight ACCESS requests on the image.
We use the same official all-parameters ACCESS checkpoint \citep{martin-etal-2020-controllable} at every size, with its NIST/SentencePiece preprocessing and fairseq beam search (beam 8, length penalty 1, maximum decoding length 200).
The 10,000-piece SentencePiece model was recreated with the official code and WikiLarge training text.
Source generation runs on Apple MPS; ACCESS runs on CPU in float32.
The complete decoded text is scored, retaining native punctuation spacing and all sentences, including truncated outputs in the denominator.

Each size searches 32 control vectors per request combination on all 50 development images.
The grid combines length multipliers of 0.7, 0.9, 1.1, and 1.3; word-rank ratios of 0.6, 0.9, 1.2, and 1.4; and dependency-depth ratios of 0.6 and 1.4, with the Levenshtein control for the amount of paraphrasing fixed at 0.75.
The character-length control is the length multiplier times the target word-count midpoint divided by the source word count, rounded to a 0.05 bucket and clipped to the checkpoint's 0.05--2.0 range.
We select the highest development joint success with two-word tolerance, breaking ties by strict joint success, mean axis compliance, and candidate order.
Selections are frozen before test generation.
The 4B search initially used a 0.4--1.4 range, which clipped many short requests; it was repeated with the wider range before testing.
The 8B and 32B searches used the wider range from the start.

Table~\ref{tab:access-details} reports strict joint success and the paired SFT+DPO gains over ACCESS with two-word tolerance, using the same image-bootstrap procedure as Table~\ref{tab:main}.
ACCESS's native word-frequency, dependency-depth, and character-length controls are mapped to our vocabulary-list, clause-presence, and word-count criteria through development selection.
These comparisons evaluate the complete generation procedures, including the source caption and rewriting stage; decoding budgets differ from direct generation.

\begin{table*}[t]
  \centering
  \small
  \input{tables/teacher_access}
  \caption{ACCESS strict joint success (\%) and joint-success gains of SFT+DPO over the ACCESS pipeline (pp) with two-word tolerance. Source size is the Qwen caption generator's size. Each set contains 100 images and 800 outputs per method. Brackets are 95\% image-bootstrap intervals, paired for differences.}
  \label{tab:access-details}
\end{table*}

The comparisons above use one uncontrolled caption per image as the rewrite source. Table~\ref{tab:access-regimes} varies that source over the same 200 images (sets 1 and 2 pooled, strict scoring): saved or freshly generated uncontrolled captions, captions constrained to the advanced--complex--long condition, and the direct output for each requested condition used as its own source. No variant exceeds 23.6\%, against 47.1\% for direct 4B SFT, so within the caption-then-rewrite pipelines we evaluated, the gap to direct generation does not close when the source caption is better matched to the requests. ACCESS controls were selected for the uncontrolled 4B source and were not retuned for the other sources.

\begin{table}[t]
  \centering
  \small
  \setlength{\tabcolsep}{3pt}
  \input{tables/access_regimes}
  \caption{ACCESS strict joint success (\%) by source caption, pooled over sets 1 and 2 (1,600 outputs per row). The direct SFT row involves no rewriting and is shown as a reference.}
  \label{tab:access-regimes}
\end{table}

\subsection{CaptionSmiths training and evaluation}
\label{app:captionsmiths}

We initialize the official CaptionSmiths implementation from LLaVA-1.5-7B \citep{liu-etal-2024-improved} and train it on the 2,502 original COCO reference captions for our 500 training images.
The condition features follow CaptionSmiths: token length, descriptiveness, and word uniqueness, normalized using the training captions.
One warmup epoch (157 updates) trains the condition embeddings and image projector; the second stage adds LoRA updates (rank 16, scaling factor 32) for five complete passes over all 2,502 captions, giving 3,130 optimizer updates at batch size 4.
Warmup uses effective batch size 16, with learning rates $10^{-3}$ for condition embeddings and $10^{-5}$ for the projector; the second stage uses $2\times10^{-5}$.
We merge the adapters and verify logits after export and reload.
This locally trained implementation uses the same training images as our students, with COCO reference captions as supervision in place of teacher-generated targets.

For each variant and request combination, we select one of eight control vectors on all 50 development images.
The grid combines normalized length values 0.150 and 0.388 with descriptiveness and uniqueness values of 0.2 and 0.8 each.
The length values come from a word-to-token regression fitted on the training captions, evaluated at the requested word-count midpoints.
Selection maximizes joint success with two-word tolerance, breaking ties by strict joint success, mean axis compliance, and candidate ID.
The native variant uses the question ``What do you see happening in this image?''; the task-instruction variant adds our vocabulary, syntax, length, and single-sentence request.
Each variant freezes its separately selected controls before testing.
Both use greedy decoding with a 96-token limit, repetition penalty 1.2, bfloat16 weights, and batch size 16 on an A100 GPU.
We score all 800 outputs per set using the main evaluation rules, including format in joint success.

Table~\ref{tab:captionsmiths-details} reports strict joint success and paired differences from the 4B SFT+DPO student.
In a post-evaluation diagnostic on ten development images, increasing the native length control from 0 to 1 raises median output length from 8 to 17 words, with descriptiveness and uniqueness fixed at 0.2.
On set 1, the native variant satisfies 48\% of long-length requests, while advanced-vocabulary and complex-syntax pass rates range from 6--9\% and 13--19\% across the corresponding conditions.
The locally trained model therefore responds to length inputs, with limited compliance for advanced vocabulary and complex syntax.

An earlier run used eight warmup updates and one epoch of condition training, reaching 11.4\% and 11.1\% joint success on the same test sets.
We increased both training budgets after inspecting those results and fixed evaluation at the fifth condition-training epoch before starting the new run.
The comparison reflects changes to both warmup and condition training; the development-control grid and evaluation rules are shared.

\begin{table*}[t]
  \centering
  \small
  \input{tables/teacher_captionsmiths}
  \caption{CaptionSmiths strict joint success (\%) and 4B SFT+DPO gains over each variant (pp) with two-word tolerance. Each set contains 100 images and 800 outputs per method. Brackets are 95\% image-bootstrap intervals, paired for differences.}
  \label{tab:captionsmiths-details}
\end{table*}

\subsection{Additional 4B comparisons}
\label{app:revision-baselines}

Table~\ref{tab:teacher-baselines} compares the distilled 4B model with additional prompts and candidate selection on the same first-set inputs.
Under two-word tolerance, its difference from four-candidate selection is uncertain (+1.375 pp [$-2.000$, $+4.750$]).
With strict selection and evaluation, joint success is 48.5\% for SFT+DPO and 43.875\% for selection, a paired difference of +4.625 pp [$+0.122$, $+9.125$].
This comparison depends on the length rule and does not measure inference latency.

\begin{table*}[t]
  \centering
  \small
  \input{tables/teacher_prompt_baselines}
  \caption{Distilled 4B students versus additional prompts and candidate selection on the first image set (800 outputs per method). Rates are percentages; differences are percentage points. Trained methods use 125B-teacher descriptions. Selection and evaluation use the same length rule within each column. Candidate budgets exclude training and rule evaluation.}
  \label{tab:teacher-baselines}
\end{table*}

\paragraph{Auxiliary comparison and generation settings.}
The trained methods in Table~\ref{tab:revision-baselines} use 4B-generated descriptions and the shared preference pairs described in Appendix~\ref{app:small-teacher-method}. The 4B prompt-only model samples four candidates per image--condition input on the first image set (3,200 candidates, temperature 0.7, top-$p$ 0.9, no top-$k$ restriction, 96-token limit). For each candidate budget $k\in\{1,2,4\}$, we retain the first candidate among the first $k$ that passes vocabulary, syntax, length, and format; if none passes, we retain candidate zero. The ordering is the saved sampling order. Each method therefore returns one output for every input, including failures; no difficult inputs are discarded.

We match the selector's length rule to the evaluation rule, using the same candidate pool for both. The original selector, which uses strict length ranges, scores 51.4\% under tolerant evaluation, whereas a selector using the two-word tolerance scores 59.1\%. Reselection changes 129 of 800 outputs. The quality ratings for that strict selector below therefore do not describe the tolerance-based selector.

\begin{table*}[t]
  \centering
  \small
  \input{tables/revision_baselines}
  \caption{4B joint success (\%) on the first 100-image set (800 outputs per method). Trained methods use 4B-generated descriptions. Selection and evaluation use the same length rule in each column. All methods retain the same image--condition inputs. ``Orig.'' denotes DPO with original pairs. Candidate budgets exclude training and rule-evaluation costs; they are not latency measurements. Paired intervals for the main contrasts are reported in the text. Bold and underlining mark the best and second-best rates in each column.}
  \label{tab:revision-baselines}
\end{table*}

At $k=1$, the first sampled candidate has 37.2\% tolerant success versus 36.0\% for greedy prompting; sampling alone is not the same intervention as selection. At $k=4$, rule-based selection exceeds the 4B-generated-data SFT+DPO model under both length rules: 59.1\% versus 53.8\% with tolerance (+5.4 pp [$+1.8$, $+9.0$]), and 43.9\% versus 39.6\% strictly (+4.3 pp [$+0.4$, $+8.0$]). All candidate-budget analyses use one saved sampling run and are exploratory; intervals measure image variation, not sampling-seed variation.

\paragraph{Rule-disclosed and two-shot prompts.}
Two further 4B baselines use the same first-set inputs. The first appends the evaluator's vocabulary thresholds, lexical-coverage requirement, clause criteria, word-count convention, and format checks to the original prompt. The second uses that same query and two image--description demonstrations from the 4B-generated SFT training data. For each condition, demonstrations come from the first two distinct training image identifiers in ascending order; all 16 descriptions pass the strict rules. Each demonstration includes its image, original prompt, and generated answer as a preceding dialogue turn. No demonstration comes from development or evaluation images.

These prompts and the example-selection rule were fixed once before their outputs were generated, without tuning them on those outputs. Greedy decoding, the pinned base weights, the 96-token limit, and image resizing match the original experiment; the new batch size is eight instead of the archived 4B command's 16. Numerical effects of this batching difference were not isolated. The two-shot query processes three images, so its single output-generation step does not have the same input cost as zero-shot prompting. These comparisons are confined to 4B and the first image set; they do not establish the performance of stronger prompts at 8B/32B or on the second set.

Both runs cover all 800 inputs, with no generation reaching the token limit. Explicit rules give 31.5\% joint success with tolerance and 21.5\% strictly; adding the two demonstrations gives 29.4\% and 20.9\%. SFT+DPO trained on 4B-generated descriptions exceeds these prompts by 22.3 pp [$+18.6$, $+25.8$] and 24.4 pp [$+20.4$, $+28.5$], respectively, under the primary rule. The original greedy prompt remains the best of these three fixed single-generation prompts. The result does not show that providing rules or examples generally harms performance: only one rule wording and one deterministic demonstration choice were tested.

\subsection{Automatic quality proxies}
\label{app:revision-quality}

This evaluation covers students trained on 4B-generated descriptions and prompt-only baselines; none of the 125B-teacher students was rated. A Qwen3-VL-32B-Instruct judge scored six 4B methods on a fixed 30-image subset of the first set (eight conditions, 240 outputs per method). Each method covers identical image--condition keys, and each evaluated sentence was matched to its saved generation. The judge sees the image and sentence for unsupported content, and the sentence alone for naturalness. It answers separate Yes/No questions; normalized next-token scores above 0.5 for unsupported content or below 0.5 for naturalness define the flags in Table~\ref{tab:revision-quality}. These scores are not calibrated probabilities of human judgments. The judge generated none of these 4B outputs but belongs to the same model family.

The unsupported-content question is: ``Does the sentence state anything that cannot be seen in the photo, such as objects, actions, reasons, feelings, intentions or events that are not visible? Answer Yes or No.'' The naturalness question is: ``Is this a natural, grammatical English sentence that a fluent writer could produce? Ignore whether it is true. Answer Yes or No.'' Each follows the sentence being assessed. The script fixes judge revision \nolinkurl{0cfaf48183f594c314753d30a4c4974bc75f3ccb}; saved quality rows identify the model but do not record the revision individually. The subset was drawn with seed 2026, and no human ratings were substituted for model scores.

The two-stage comparator first asks the unadapted 4B model for one natural, faithful sentence about each image without linguistic targets. The same model then rewrites that sentence under each of the eight original conditions, using the description text without seeing the image again. Both stages use greedy decoding and a 96-token limit. One first-stage description is shared across conditions, giving 100 descriptions plus 800 rewrites on the full first set; Table~\ref{tab:revision-quality} rates only the same 30-image subset used for the other methods.

\begin{table*}[t]
  \centering
  \small
  \input{tables/revision_quality}
  \caption{Automatic quality flags for the 4B-generated-data experiment, on the same 30 images and eight conditions (240 outputs per method). Lower is better for both columns; bold and underlining mark the best and second-best rates. Brackets are 95\% image-bootstrap intervals. ``Orig.'' denotes DPO with original pairs. Best-of-4 here uses strict selection; the two-stage baseline describes the image and then rewrites that description. These are model flags, not human-rated error rates.}
  \label{tab:revision-quality}
\end{table*}

The 4B-generated-data SFT+DPO model increases unsupported-content flags from 10.8\% to 16.7\%, a difference of 5.8 pp [$+1.7$, $+10.0$] relative to prompting. Unnaturalness flags rise from 2.9\% to 3.8\%, with an uncertain difference (+0.8 pp [$-2.1$, $+3.8$]); this does not establish equivalence. A separate Claude visual audit (model version unrecorded) examined 18 complex-request examples (12 flagged, six unflagged): it disagreed with six flags and found no unsupported content in the six unflagged examples. That audit is another model judgment, not a human validation, and cannot calibrate these rates. The flags warrant caution about faithfulness. No quality rating is inferred for unassessed 8B/32B outputs, tolerance-based selections, or the new rule-disclosed and two-shot prompt outputs.

\subsection{Condition-level results with 4B-generated targets}

Tables~\ref{tab:smallteacher-conditions1} and \ref{tab:smallteacher-conditions2} report each vocabulary--syntax--length combination for the 4B-generated-data students. As in Tables~\ref{tab:conditions1} and \ref{tab:conditions2}, each cell uses 100 outputs, one per image.

\begin{table*}[t]
  \centering
  \small
  \setlength{\tabcolsep}{3pt}
  \input{tables/short_conditions}
  \caption{Joint success (\%) for students trained on 4B-generated descriptions, by condition on the first image set, with two-word length tolerance. P: prompt only; S: SFT; O: DPO with original pairs; D: SFT+DPO. There are 100 outputs per cell.}
  \label{tab:smallteacher-conditions1}
\end{table*}

\begin{table*}[t]
  \centering
  \small
  \setlength{\tabcolsep}{3pt}
  \input{tables/short_conditions_test2}
  \caption{The same 4B-generated-data comparison on the second image set (100 outputs per cell). The abbreviations and two-word tolerance are as in Table~\ref{tab:smallteacher-conditions1}.}
  \label{tab:smallteacher-conditions2}
\end{table*}

\subsection{Examples of length violations}

Table~\ref{tab:examples} shows two 32B comparisons from the 4B-generated-data experiment for the same image and instruction. Both request a short sentence, so the main metric accepts at most 16 words. DPO with original pairs adds words while preserving the requested vocabulary and clause criterion; SFT+DPO returns within the accepted range. These examples illustrate word-count behavior, not image faithfulness. For example, describing a bus as driving, waiting, or parked can change the claim about the image even when every linguistic check passes. No human judgment verifies those descriptions.

\begin{table*}[t]
  \centering
  \small
  \input{tables/examples_32b}
  \caption{Selected 32B outputs from the 4B-generated-data experiment for identical image--condition inputs. ``Orig.'' denotes DPO with original pairs. All three methods pass the vocabulary and syntax checks; DPO with original pairs exceeds the accepted short-sentence limit. Selection takes the first two matches among six fixed example images and then other images ordered by identifier. These are illustrative cases, not an estimate of frequency.}
  \label{tab:examples}
\end{table*}

\subsection{Teacher generation reference}

Table~\ref{tab:teacher} reports the 125B teacher's greedy generation alongside the distilled students on the same image sets.
The teacher uses Q5\_K\_M quantization through llama.cpp, as in target generation.
Differences in architecture and inference software prevent treating it as another point in the student-size comparison.

\begin{table*}[t]
  \centering
  \small
  \setlength{\tabcolsep}{3pt}
  \input{tables/short_teacher}
  \caption{Teacher and distilled-student joint success (\%) on the same two image sets, with 800 outputs per cell and 95\% image-bootstrap intervals. SFT uses filtered 125B-teacher descriptions; DPO uses each student's own candidates.}
  \label{tab:teacher}
\end{table*}

\subsection{Faithfulness screen audit}
\label{app:screen-audit}

The SFT data pipeline applies a faithfulness screen (the unadapted 4B model judging its teacher's candidates) after the vocabulary, grammar, length, and format rules. We audited this screen with an independent judge, Gemma-4-31B, which labelled the atomic visual claims of candidate sentences against the image as supported, contradicted, or undetermined; a sentence counts as contradicted when any claim is contradicted. Of 21,828 teacher candidates, 5,828 pass all rules; the screen removes 245 of them (4.2\%) and retains 5,583 (95.8\%). All 245 removed candidates and a weighted sample of 300 retained candidates were judged. Among the removed candidates, 30.2\% [23.9, 37.0] were contradicted, 46.5\% [38.5, 53.9] supported, and 23.3\% undetermined; among retained candidates, 8.5\% [5.5, 11.7] were contradicted. The screen therefore enriches contradictions among the removed candidates by about 3.5 times while discarding some supported sentences: it is a conservative screen, not a precise filter, and the retained data still contain contradicted sentences. As a separate check on evaluation images, the same screen rejected 19 of 522 rule-passing teacher captions (3.6\%); 9 of the 19 were judged supported. These rates come from an automated judge and are not human error probabilities.

\subsection{Independent visual audit}
\label{app:visual-audit}

We audited the visual content of the 4B outputs with the same independent Gemma-4-31B judge, against fact inventories fixed before scoring (30 confirmation images $\times$ eight requests $\times$ three methods; 240 outputs per method). Table~\ref{tab:visual-audit} reports the share of atomic claims the judge marks supported (precision), the share of inventoried salient facts the output mentions (recall), and the share of outputs containing at least one contradicted claim (unsupported). Precision is high for all three methods. SFT lowers the unsupported-output rate from 20.8\% to 14.6\% overall, with the clearest gain on long requests (27.5\% to 15.0\%; paired difference $-$12.5 points [$-$24.2, $-$2.5]) and a small recall gain there (+3.4 points [+0.3, +6.7]). Under short requests, recall is nearly identical for prompting and SFT (26.3\% and 26.1\%), so the compression to 10--14 words does not detectably reduce coverage of the inventoried facts relative to prompting at the same lengths; both methods cover less than at long lengths, which is the expected effect of the shorter budget. The judge is a single automated model, so these labels complement, and do not replace, the human faithfulness ratings of Section~\ref{sec:human-quality}.

\begin{table*}[t]
  \centering
  \small
  \setlength{\tabcolsep}{3pt}
  \input{tables/visual_audit}
  \caption{Independent visual audit of 4B outputs (\%, 240 outputs per method). Precision is the mean per-output share of supported atomic claims; the unsupported rate is the share of outputs with at least one contradicted claim; recall is against fact inventories fixed before scoring. Long requests are the 20--24-word conditions (120 outputs per method).}
  \label{tab:visual-audit}
\end{table*}

\subsection{Oracle candidate selection and latency}
\label{app:oracle}

Table~\ref{tab:oracle} compares single-pass SFT with prompt-only candidate selection on the confirmation set under strict scoring. The selector is the evaluation rule set itself, so best-of-$k$ prompting is an oracle upper bound for selection-based inference. Best-of-four prompting remains below single-pass SFT, and its batched latency exceeds that of a single SFT pass. Merging the LoRA adapter into the base weights lowers median latency from 1.69\,s to 1.36\,s per output, with a small and uncertain accuracy change ($-$1.1 points [$-$2.6, +0.3]). Latency was measured on one Apple Silicon device (MPS) over 20 images $\times$ eight requests $\times$ two repeats; the candidates for the latency runs are separate draws from those scored for accuracy.

\begin{table}[t]
  \centering
  \small
  \setlength{\tabcolsep}{3pt}
  \input{tables/oracle_latency}
  \caption{Oracle selection and latency on the confirmation set (800 outputs per method). Differences are against single-pass SFT, in percentage points, with 95\% paired image-bootstrap intervals.}
  \label{tab:oracle}
\end{table}

\subsection{Output diversity}
\label{app:diversity}

Table~\ref{tab:diversity} summarises output variability on the confirmation set (7,200 outputs). Training does not collapse the outputs onto a small set of templates: the share of outputs beginning with the most common three-gram stays near 3--5\%, bigram repetition stays near 3\%, and at 32B SFT reduces repetition from 7.5\% to 2.9\% while raising distinct-1 from 0.335 to 0.351. Pairwise word overlap rises slightly at 4B (0.076 to 0.091), alongside somewhat longer outputs (17.6 versus 16.0 words on average).

\begin{table*}[t]
  \centering
  \small
  \setlength{\tabcolsep}{3pt}
  \input{tables/diversity}
  \caption{Output variability on the confirmation set, averaged over the eight conditions (100 outputs per condition). ``Opening'' is the share of outputs starting with the most common opening three-gram; ``Jaccard'' is mean pairwise word overlap within a condition; ``Rep.\ bigram'' is the repeated-bigram rate.}
  \label{tab:diversity}
\end{table*}

%% file: tables/teacher_sensitivity.tex
\begin{tabular}{llcccc}
\toprule
Size & Method & Set 1, strict & Set 1, $\pm2$ & Set 2, strict & Set 2, $\pm2$ \\
\midrule
4B & Prompt & \underline{22.6} \ci{19.8}{25.6} & 36.0 \ci{32.9}{39.2} & 20.9 \ci{18.1}{23.6} & 35.4 \ci{32.0}{38.8} \\
4B & SFT & \textbf{48.5} \ci{44.9}{52.2} & \underline{59.4} \ci{55.9}{62.9} & \underline{45.8} \ci{42.5}{49.1} & \underline{55.6} \ci{52.2}{59.0} \\
4B & SFT+DPO & \textbf{48.5} \ci{44.8}{52.2} & \textbf{60.5} \ci{57.0}{64.1} & \textbf{46.8} \ci{43.2}{50.2} & \textbf{57.2} \ci{53.9}{60.6} \\
8B & Prompt & 24.9 \ci{22.4}{27.4} & 38.6 \ci{35.9}{41.4} & 25.6 \ci{22.8}{28.5} & 38.0 \ci{34.8}{41.4} \\
8B & SFT & \underline{47.8} \ci{44.4}{51.2} & \underline{57.4} \ci{53.9}{61.0} & \underline{43.6} \ci{40.4}{46.9} & \underline{53.6} \ci{50.4}{56.9} \\
8B & SFT+DPO & \textbf{51.9} \ci{48.2}{55.4} & \textbf{60.5} \ci{56.8}{64.2} & \textbf{52.0} \ci{48.1}{55.8} & \textbf{60.5} \ci{56.9}{64.0} \\
32B & Prompt & 17.1 \ci{14.2}{20.1} & 31.8 \ci{28.1}{35.5} & 17.6 \ci{15.0}{20.4} & 32.8 \ci{29.4}{36.1} \\
32B & SFT & \textbf{51.5} \ci{47.8}{55.2} & \underline{59.6} \ci{56.2}{62.9} & \underline{49.5} \ci{46.0}{53.0} & \underline{57.6} \ci{54.4}{61.0} \\
32B & SFT+DPO & \underline{51.4} \ci{47.8}{55.1} & \textbf{61.6} \ci{58.4}{64.9} & \textbf{52.2} \ci{48.8}{55.9} & \textbf{59.8} \ci{56.4}{63.2} \\
\bottomrule
\end{tabular}

%% file: tables/teacher_conditions.tex
\begin{tabular}{lccccccccc}
\toprule
Condition & 4B P & 4B S & 4B D & 8B P & 8B S & 8B D & 32B P & 32B S & 32B D \\
\midrule
basic-simple-short & 81.0 & \textbf{85.0} & \underline{83.0} & 82.0 & \underline{83.0} & \textbf{84.0} & 61.0 & \textbf{85.0} & \underline{84.0} \\
basic-simple-long & 41.0 & \underline{78.0} & \textbf{79.0} & \underline{34.0} & \textbf{76.0} & \textbf{76.0} & 61.0 & \underline{70.0} & \textbf{75.0} \\
basic-complex-short & 19.0 & \underline{75.0} & \textbf{84.0} & 51.0 & \underline{75.0} & \textbf{84.0} & 23.0 & \underline{77.0} & \textbf{81.0} \\
basic-complex-long & 43.0 & \textbf{79.0} & \underline{76.0} & 67.0 & \underline{69.0} & \textbf{78.0} & 47.0 & \underline{77.0} & \textbf{82.0} \\
advanced-simple-short & 28.0 & \underline{40.0} & \textbf{41.0} & 18.0 & \underline{34.0} & \textbf{40.0} & 18.0 & \underline{34.0} & \textbf{36.0} \\
advanced-simple-long & 24.0 & \underline{37.0} & \textbf{38.0} & 10.0 & \underline{42.0} & \textbf{46.0} & 24.0 & \underline{45.0} & \textbf{47.0} \\
advanced-complex-short & 14.0 & \underline{36.0} & \textbf{38.0} & 17.0 & \textbf{34.0} & \underline{32.0} & \underline{5.0} & \textbf{38.0} & \textbf{38.0} \\
advanced-complex-long & \underline{38.0} & \textbf{45.0} & \textbf{45.0} & 30.0 & \textbf{46.0} & \underline{44.0} & 15.0 & \textbf{51.0} & \underline{50.0} \\
\bottomrule
\end{tabular}

%% file: tables/teacher_conditions_test2.tex
\begin{tabular}{lccccccccc}
\toprule
Condition & 4B P & 4B S & 4B D & 8B P & 8B S & 8B D & 32B P & 32B S & 32B D \\
\midrule
basic-simple-short & 69.0 & \underline{80.0} & \textbf{81.0} & \underline{78.0} & \underline{78.0} & \textbf{79.0} & 67.0 & \underline{77.0} & \textbf{83.0} \\
basic-simple-long & 41.0 & \textbf{65.0} & \underline{63.0} & 36.0 & \underline{69.0} & \textbf{76.0} & 54.0 & \underline{72.0} & \textbf{77.0} \\
basic-complex-short & 13.0 & \underline{68.0} & \textbf{77.0} & 49.0 & \underline{68.0} & \textbf{80.0} & 23.0 & \textbf{80.0} & \underline{79.0} \\
basic-complex-long & 35.0 & \underline{68.0} & \textbf{73.0} & \underline{65.0} & 62.0 & \textbf{75.0} & \underline{48.0} & \textbf{74.0} & \textbf{74.0} \\
advanced-simple-short & \underline{35.0} & \underline{35.0} & \textbf{37.0} & \underline{19.0} & \textbf{36.0} & \textbf{36.0} & 18.0 & \underline{32.0} & \textbf{35.0} \\
advanced-simple-long & 30.0 & \underline{43.0} & \textbf{47.0} & 12.0 & \underline{45.0} & \textbf{51.0} & 31.0 & \underline{45.0} & \textbf{48.0} \\
advanced-complex-short & 23.0 & \textbf{34.0} & \underline{32.0} & 16.0 & \underline{31.0} & \textbf{37.0} & 3.0 & \textbf{38.0} & \underline{36.0} \\
advanced-complex-long & 37.0 & \textbf{52.0} & \underline{48.0} & 29.0 & \underline{40.0} & \textbf{50.0} & 18.0 & \underline{43.0} & \textbf{46.0} \\
\bottomrule
\end{tabular}

%% file: tables/teacher_axis_conditions.tex
\begin{tabular}{lccccccccc}
\toprule
Condition & 4B V & 4B G & 4B L & 8B V & 8B G & 8B L & 32B V & 32B G & 32B L \\
\midrule
basic-simple-short & 84.0 & \textbf{99.0} & \textbf{100.0} & \underline{85.0} & \underline{98.0} & \textbf{100.0} & \textbf{87.0} & 97.0 & \textbf{100.0} \\
basic-simple-long & \textbf{91.0} & \underline{88.0} & \underline{97.0} & 82.0 & \underline{88.0} & \textbf{98.0} & \underline{86.0} & \textbf{90.0} & \textbf{98.0} \\
basic-complex-short & \textbf{87.0} & \textbf{100.0} & \underline{97.0} & \textbf{87.0} & \textbf{100.0} & \textbf{98.0} & \underline{85.0} & \textbf{100.0} & \underline{97.0} \\
basic-complex-long & 78.0 & \textbf{100.0} & \underline{97.0} & \underline{81.0} & \textbf{100.0} & \textbf{98.0} & \textbf{85.0} & \textbf{100.0} & 96.0 \\
advanced-simple-short & \textbf{42.0} & 97.0 & \underline{99.0} & \underline{40.0} & \textbf{100.0} & \textbf{100.0} & 37.0 & \underline{99.0} & \underline{99.0} \\
advanced-simple-long & 46.0 & 89.0 & 90.0 & \underline{48.0} & \textbf{97.0} & \textbf{99.0} & \textbf{49.0} & \underline{96.0} & \underline{98.0} \\
advanced-complex-short & \textbf{42.0} & \textbf{100.0} & 95.0 & 33.0 & \textbf{100.0} & \textbf{97.0} & \underline{39.0} & \underline{99.0} & \underline{96.0} \\
advanced-complex-long & \underline{48.0} & \textbf{100.0} & 90.0 & 46.0 & \textbf{100.0} & \textbf{98.0} & \textbf{53.0} & \underline{99.0} & \underline{94.0} \\
\bottomrule
\end{tabular}

%% file: tables/teacher_axis_conditions_test2.tex
\begin{tabular}{lccccccccc}
\toprule
Condition & 4B V & 4B G & 4B L & 8B V & 8B G & 8B L & 32B V & 32B G & 32B L \\
\midrule
basic-simple-short & \textbf{84.0} & 97.0 & \underline{98.0} & 80.0 & \underline{99.0} & \textbf{100.0} & \underline{83.0} & \textbf{100.0} & \textbf{100.0} \\
basic-simple-long & 81.0 & 86.0 & 92.0 & \underline{84.0} & \textbf{96.0} & \underline{95.0} & \textbf{86.0} & \underline{91.0} & \textbf{100.0} \\
basic-complex-short & \underline{80.0} & \textbf{100.0} & 97.0 & \underline{80.0} & \textbf{100.0} & \textbf{100.0} & \textbf{81.0} & \textbf{100.0} & \underline{98.0} \\
basic-complex-long & 76.0 & \textbf{100.0} & 96.0 & \underline{77.0} & \underline{99.0} & \textbf{99.0} & \textbf{78.0} & \underline{99.0} & \underline{97.0} \\
advanced-simple-short & \textbf{39.0} & \underline{99.0} & 98.0 & 36.0 & \textbf{100.0} & \textbf{100.0} & \underline{37.0} & 97.0 & \underline{99.0} \\
advanced-simple-long & 53.0 & \underline{90.0} & \underline{94.0} & \textbf{57.0} & \textbf{92.0} & \textbf{98.0} & \underline{55.0} & \underline{90.0} & \textbf{98.0} \\
advanced-complex-short & \underline{36.0} & \textbf{100.0} & 93.0 & \textbf{37.0} & \textbf{100.0} & \textbf{98.0} & \textbf{37.0} & \textbf{100.0} & \underline{95.0} \\
advanced-complex-long & \textbf{53.0} & \textbf{100.0} & 87.0 & \underline{52.0} & \textbf{100.0} & \textbf{97.0} & 48.0 & \underline{99.0} & \underline{92.0} \\
\bottomrule
\end{tabular}

%% file: tables/short_main.tex
\begin{tabular}{llccccc}
\toprule
Size & Method & Vocabulary & Syntax & Length & Joint (set 1) & Joint (set 2) \\
\midrule
4B & Prompt & 60.1 & 74.4 & 87.2 & 36.0 \ci{32.9}{39.2} & 35.4 \ci{32.0}{38.8} \\
4B & SFT & 60.8 & 82.6 & \textbf{91.4} & 44.9 \ci{41.2}{48.5} & 43.2 \ci{40.0}{46.5} \\
4B & SFT+DPO (orig.) & \underline{60.9} & \underline{93.1} & \underline{91.0} & \underline{51.0} \ci{47.5}{54.6} & \underline{52.8} \ci{49.2}{56.2} \\
4B & SFT+DPO & \textbf{63.6} & \textbf{93.9} & 90.4 & \textbf{53.8} \ci{50.1}{57.5} & \textbf{52.9} \ci{49.2}{56.4} \\
8B & Prompt & 55.2 & 92.0 & 76.6 & 38.6 \ci{35.9}{41.4} & 38.0 \ci{34.8}{41.4} \\
8B & SFT & 62.5 & 87.5 & \textbf{95.0} & 50.5 \ci{47.0}{54.1} & 45.4 \ci{42.0}{48.9} \\
8B & SFT+DPO (orig.) & \underline{66.4} & \underline{95.1} & 86.0 & \underline{54.5} \ci{51.0}{58.0} & \underline{49.0} \ci{45.8}{52.2} \\
8B & SFT+DPO & \textbf{70.4} & \textbf{95.2} & \underline{92.6} & \textbf{61.1} \ci{57.5}{64.6} & \textbf{55.0} \ci{51.5}{58.5} \\
32B & Prompt & 57.6 & 95.1 & 53.1 & 31.8 \ci{28.1}{35.5} & 32.8 \ci{29.4}{36.1} \\
32B & SFT & 63.9 & 90.1 & \underline{96.0} & 54.5 \ci{51.2}{57.8} & 50.1 \ci{46.5}{53.8} \\
32B & SFT+DPO (orig.) & \textbf{66.5} & \underline{95.6} & 86.2 & \underline{55.0} \ci{50.9}{59.1} & \underline{51.2} \ci{47.8}{54.8} \\
32B & SFT+DPO & \underline{65.9} & \textbf{96.5} & \textbf{97.4} & \textbf{61.5} \ci{57.9}{65.0} & \textbf{59.2} \ci{55.9}{62.6} \\
\bottomrule
\end{tabular}

%% file: tables/short_sensitivity.tex
\begin{tabular}{llcccc}
\toprule
Size & Method & Set 1, strict & Set 1, $\pm2$ & Set 2, strict & Set 2, $\pm2$ \\
\midrule
4B & Prompt & 22.6 \ci{19.8}{25.6} & 36.0 \ci{32.9}{39.2} & 20.9 \ci{18.1}{23.6} & 35.4 \ci{32.0}{38.8} \\
4B & SFT & 30.8 \ci{27.9}{33.6} & 44.9 \ci{41.2}{48.5} & 29.9 \ci{27.0}{32.8} & 43.2 \ci{40.0}{46.5} \\
4B & SFT+DPO (orig.) & \underline{36.6} \ci{33.4}{39.9} & \underline{51.0} \ci{47.5}{54.6} & \underline{37.0} \ci{33.8}{40.2} & \underline{52.8} \ci{49.2}{56.2} \\
4B & SFT+DPO & \textbf{39.6} \ci{36.0}{43.2} & \textbf{53.8} \ci{50.1}{57.5} & \textbf{38.8} \ci{35.6}{41.9} & \textbf{52.9} \ci{49.2}{56.4} \\
8B & Prompt & 24.9 \ci{22.4}{27.4} & 38.6 \ci{35.9}{41.4} & 25.6 \ci{22.8}{28.5} & 38.0 \ci{34.8}{41.4} \\
8B & SFT & \underline{37.6} \ci{34.4}{41.0} & 50.5 \ci{47.0}{54.1} & 33.0 \ci{29.6}{36.5} & 45.4 \ci{42.0}{48.9} \\
8B & SFT+DPO (orig.) & 36.5 \ci{32.8}{40.2} & \underline{54.5} \ci{51.0}{58.0} & \underline{33.6} \ci{30.2}{37.0} & \underline{49.0} \ci{45.8}{52.2} \\
8B & SFT+DPO & \textbf{46.9} \ci{43.1}{50.5} & \textbf{61.1} \ci{57.5}{64.6} & \textbf{41.8} \ci{38.0}{45.4} & \textbf{55.0} \ci{51.5}{58.5} \\
32B & Prompt & 17.1 \ci{14.2}{20.1} & 31.8 \ci{28.1}{35.5} & 17.6 \ci{15.0}{20.4} & 32.8 \ci{29.4}{36.1} \\
32B & SFT & \underline{43.0} \ci{39.9}{46.2} & 54.5 \ci{51.2}{57.8} & \underline{36.6} \ci{33.1}{40.1} & 50.1 \ci{46.5}{53.8} \\
32B & SFT+DPO (orig.) & 36.5 \ci{32.6}{40.4} & \underline{55.0} \ci{50.9}{59.1} & 33.8 \ci{30.5}{37.1} & \underline{51.2} \ci{47.8}{54.8} \\
32B & SFT+DPO & \textbf{49.4} \ci{46.0}{52.8} & \textbf{61.5} \ci{57.9}{65.0} & \textbf{46.9} \ci{43.2}{50.5} & \textbf{59.2} \ci{55.9}{62.6} \\
\bottomrule
\end{tabular}

%% file: tables/teacher_access.tex
\begin{tabular}{lcccc}
\toprule
Source size & Strict joint (set 1) & Strict joint (set 2) & SFT+DPO gain (set 1) & SFT+DPO gain (set 2) \\
\midrule
4B & 17.6 \ci{15.5}{19.6} & 17.8 \ci{15.8}{19.9} & +36.5 \ci{+32.6}{+40.6} & +34.0 \ci{+30.4}{+37.8} \\
8B & 16.1 \ci{14.2}{18.0} & 17.4 \ci{15.6}{19.0} & +38.9 \ci{+34.4}{+43.5} & +38.1 \ci{+33.9}{+42.4} \\
32B & 17.4 \ci{15.2}{19.5} & 16.4 \ci{14.2}{18.5} & +39.4 \ci{+35.2}{+43.6} & +37.9 \ci{+33.9}{+41.9} \\
\bottomrule
\end{tabular}

%% file: tables/access_regimes.tex
\begin{tabular}{lc}
\toprule
ACCESS source caption & Strict joint (\%) \\
\midrule
Uncontrolled, 125B model & 17.2 \\
Uncontrolled, 4B model (saved) & 17.7 \\
Constrained 4B (adv., complex, long) & 18.0 \\
Constrained 125B (adv., complex, long) & 18.4 \\
Condition-specific, 4B model & 19.6 \\
Condition-specific, 125B model & 23.6 \\
\midrule
Direct 4B SFT (reference) & 47.1 \\
\bottomrule
\end{tabular}

%% file: tables/teacher_captionsmiths.tex
\begin{tabular}{lcccc}
\toprule
CaptionSmiths variant & Strict joint (set 1) & Strict joint (set 2) & 4B gain (set 1) & 4B gain (set 2) \\
\midrule
Native controls & 13.9 \ci{12.2}{15.5} & 14.1 \ci{12.5}{15.9} & +42.5 \ci{+38.4}{+46.8} & +39.5 \ci{+35.6}{+43.5} \\
+ task instruction & 10.1 \ci{9.0}{11.1} & 10.0 \ci{8.8}{11.2} & +49.0 \ci{+45.1}{+52.9} & +45.6 \ci{+42.1}{+49.1} \\
\bottomrule
\end{tabular}

%% file: tables/teacher_prompt_baselines.tex
\begin{tabular}{lcccc}
\toprule
4B method & $k$ & Strict & $\pm2$ & DPO minus method ($\pm2$) \\
\midrule
Prompt, greedy & 1 & 22.6 \ci{19.8}{25.6} & 36.0 \ci{32.9}{39.2} & +24.5 \ci{+20.4}{+28.6} \\
Prompt, one sample & 1 & 22.6 \ci{19.9}{25.5} & 37.2 \ci{34.1}{40.5} & +23.2 \ci{+19.0}{+27.5} \\
Prompt, best-of-2 & 2 & 34.0 \ci{30.9}{37.0} & 47.8 \ci{44.4}{51.1} & +12.8 \ci{+8.9}{+16.6} \\
Prompt, best-of-4 & 4 & \underline{43.9} \ci{40.6}{47.0} & 59.1 \ci{55.8}{62.4} & +1.4 \ci{-2.0}{+4.8} \\
Explicit rules & 1 & 21.5 \ci{18.8}{24.2} & 31.5 \ci{28.5}{34.5} & +29.0 \ci{+24.8}{+33.4} \\
Explicit rules + 2 shots & 1 & 20.9 \ci{18.5}{23.2} & 29.4 \ci{26.8}{32.1} & +31.1 \ci{+27.0}{+35.2} \\
SFT (125B targets) & 1 & \textbf{48.5} \ci{44.9}{52.2} & \underline{59.4} \ci{55.9}{62.9} & +1.1 \ci{-1.0}{+3.4} \\
SFT+DPO (125B targets) & 1 & \textbf{48.5} \ci{44.8}{52.2} & \textbf{60.5} \ci{57.0}{64.1} & --- \\
\bottomrule
\end{tabular}

%% file: tables/revision_baselines.tex
\begin{tabular}{lccc}
\toprule
4B method & $k$ & Strict & $\pm2$ \\
\midrule
Prompt, greedy & 1 & 22.6 & 36.0 \\
Prompt, one sample & 1 & 22.6 & 37.2 \\
Prompt, best-of-2 & 2 & 34.0 & 47.8 \\
Prompt, best-of-4 & 4 & \textbf{43.9} & \textbf{59.1} \\
SFT & 1 & 30.8 & 44.9 \\
SFT+DPO (orig.) & 1 & 36.6 & 51.0 \\
SFT+DPO & 1 & \underline{39.6} & \underline{53.8} \\
Explicit rules & 1 & 21.5 & 31.5 \\
Explicit rules + 2 shots & 1 & 20.9 & 29.4 \\
\bottomrule
\end{tabular}

%% file: tables/revision_quality.tex
\begin{tabular}{lrr}
\toprule
4B method & Unsupported flag (\%) & Unnatural flag (\%) \\
\midrule
Prompt & \textbf{10.8} [5.4, 17.5] & \underline{2.9} [0.4, 7.1] \\
SFT & 13.3 [6.7, 21.2] & 4.6 [1.7, 8.3] \\
SFT+DPO (orig.) & 15.4 [7.9, 24.2] & 5.0 [2.1, 8.3] \\
SFT+DPO & 16.7 [10.0, 24.6] & 3.8 [1.2, 6.7] \\
Best-of-4 (strict selector) & \underline{12.5} [6.2, 19.6] & 3.8 [0.8, 7.1] \\
Two-stage & 16.7 [7.9, 27.1] & \textbf{2.1} [0.4, 3.8] \\
\bottomrule
\end{tabular}

%% file: tables/short_conditions.tex
\begin{tabular}{lcccccccccccc}
\toprule
Condition & 4B P & 4B S & 4B O & 4B D & 8B P & 8B S & 8B O & 8B D & 32B P & 32B S & 32B O & 32B D \\
\midrule
basic-simple-short & 81.0 & \textbf{84.0} & 78.0 & \underline{82.0} & \underline{82.0} & \textbf{83.0} & 74.0 & 76.0 & 61.0 & \textbf{78.0} & \underline{76.0} & \textbf{78.0} \\
basic-simple-long & 41.0 & 53.0 & \textbf{63.0} & \underline{57.0} & 34.0 & 62.0 & \textbf{71.0} & \underline{69.0} & 61.0 & \underline{68.0} & 65.0 & \textbf{74.0} \\
basic-complex-short & 19.0 & 32.0 & \underline{51.0} & \textbf{59.0} & 51.0 & 39.0 & \underline{62.0} & \textbf{67.0} & 23.0 & \underline{58.0} & 52.0 & \textbf{67.0} \\
basic-complex-long & 43.0 & 60.0 & \underline{67.0} & \textbf{75.0} & \underline{67.0} & 65.0 & 66.0 & \textbf{74.0} & 47.0 & 63.0 & \underline{67.0} & \textbf{85.0} \\
advanced-simple-short & 28.0 & 33.0 & \underline{34.0} & \textbf{39.0} & 18.0 & 32.0 & \underline{38.0} & \textbf{43.0} & 18.0 & 40.0 & \textbf{45.0} & \underline{43.0} \\
advanced-simple-long & 24.0 & 26.0 & \textbf{34.0} & \underline{32.0} & 10.0 & 38.0 & \underline{41.0} & \textbf{51.0} & 24.0 & 40.0 & \textbf{43.0} & \underline{42.0} \\
advanced-complex-short & 14.0 & 29.0 & \underline{34.0} & \textbf{39.0} & 17.0 & 36.0 & \underline{40.0} & \textbf{50.0} & 5.0 & 33.0 & \underline{36.0} & \textbf{45.0} \\
advanced-complex-long & 38.0 & \underline{42.0} & \textbf{47.0} & \textbf{47.0} & 30.0 & \underline{49.0} & 44.0 & \textbf{59.0} & 15.0 & \underline{56.0} & \underline{56.0} & \textbf{58.0} \\
\bottomrule
\end{tabular}

%% file: tables/short_conditions_test2.tex
\begin{tabular}{lcccccccccccc}
\toprule
Condition & 4B P & 4B S & 4B O & 4B D & 8B P & 8B S & 8B O & 8B D & 32B P & 32B S & 32B O & 32B D \\
\midrule
basic-simple-short & 69.0 & \textbf{71.0} & 69.0 & \underline{70.0} & \textbf{78.0} & \underline{70.0} & 64.0 & 64.0 & 67.0 & \underline{73.0} & 69.0 & \textbf{77.0} \\
basic-simple-long & 41.0 & 49.0 & \textbf{58.0} & \underline{54.0} & 36.0 & 55.0 & \textbf{65.0} & \underline{62.0} & 54.0 & \underline{64.0} & \underline{64.0} & \textbf{71.0} \\
basic-complex-short & 13.0 & 30.0 & \underline{58.0} & \textbf{63.0} & 49.0 & 38.0 & \underline{50.0} & \textbf{66.0} & 23.0 & 44.0 & \underline{56.0} & \textbf{68.0} \\
basic-complex-long & 35.0 & 42.0 & \textbf{63.0} & \underline{62.0} & \textbf{65.0} & 47.0 & 48.0 & \underline{59.0} & 48.0 & \underline{61.0} & 56.0 & \textbf{66.0} \\
advanced-simple-short & 35.0 & 36.0 & \textbf{40.0} & \underline{37.0} & 19.0 & \textbf{35.0} & \textbf{35.0} & \underline{34.0} & 18.0 & 29.0 & \textbf{39.0} & \underline{37.0} \\
advanced-simple-long & 30.0 & \underline{37.0} & \textbf{38.0} & 35.0 & 12.0 & 29.0 & \underline{33.0} & \textbf{42.0} & 31.0 & \underline{34.0} & 33.0 & \textbf{41.0} \\
advanced-complex-short & 23.0 & 30.0 & \underline{38.0} & \textbf{42.0} & 16.0 & 39.0 & \underline{41.0} & \textbf{53.0} & 3.0 & 42.0 & \underline{48.0} & \textbf{58.0} \\
advanced-complex-long & 37.0 & 51.0 & \underline{58.0} & \textbf{60.0} & 29.0 & 50.0 & \underline{56.0} & \textbf{60.0} & 18.0 & \underline{54.0} & 45.0 & \textbf{56.0} \\
\bottomrule
\end{tabular}

%% file: tables/examples_32b.tex
\begin{tabular}{@{}l>{\raggedright\arraybackslash}p{0.76\linewidth}@{}}
\toprule
\multicolumn{2}{@{}l}{COCO image 17031; condition: advanced, complex, short} \\
SFT & A giraffe, whose large ears frame its face, peers curiously from a sunlit, leafy backdrop. (15 words) \\
SFT+DPO (orig.) & A giraffe, whose large ears frame its inquisitive gaze, peers directly at the viewer amid lush green foliage. (18 words) \\
SFT+DPO & A giraffe, whose large ears frame its face, peers curiously while foliage blurs behind it. (15 words) \\
\midrule
\multicolumn{2}{@{}l}{COCO image 88462; condition: basic, complex, short} \\
SFT & A yellow school bus, which says ``First Student,'' drives down a street with cars. (14 words) \\
SFT+DPO (orig.) & A yellow school bus, which says ``First Student,'' drives down a street while cars wait behind it. (17 words) \\
SFT+DPO & A yellow school bus, which is parked, blocks the road while cars wait behind it. (15 words) \\
\bottomrule
\end{tabular}

%% file: tables/short_teacher.tex
\begin{tabular}{llcccc}
\toprule
Size & Method & Set 1, strict & Set 1, $\pm2$ & Set 2, strict & Set 2, $\pm2$ \\
\midrule
125B & Teacher, greedy & 31.6 \ci{28.4}{34.8} & 43.2 \ci{40.2}{46.2} & 33.6 \ci{30.2}{37.2} & 45.6 \ci{41.9}{49.5} \\
4B & SFT & \textbf{48.5} \ci{44.9}{52.2} & \underline{59.4} \ci{55.9}{62.9} & \underline{45.8} \ci{42.5}{49.1} & \underline{55.6} \ci{52.2}{59.0} \\
4B & SFT+DPO & \textbf{48.5} \ci{44.8}{52.2} & \textbf{60.5} \ci{57.0}{64.1} & \textbf{46.8} \ci{43.2}{50.2} & \textbf{57.2} \ci{53.9}{60.6} \\
8B & SFT & \underline{47.8} \ci{44.4}{51.2} & \underline{57.4} \ci{53.9}{61.0} & \underline{43.6} \ci{40.4}{46.9} & \underline{53.6} \ci{50.4}{56.9} \\
8B & SFT+DPO & \textbf{51.9} \ci{48.2}{55.4} & \textbf{60.5} \ci{56.8}{64.2} & \textbf{52.0} \ci{48.1}{55.8} & \textbf{60.5} \ci{56.9}{64.0} \\
32B & SFT & \textbf{51.5} \ci{47.8}{55.2} & \underline{59.6} \ci{56.2}{62.9} & \underline{49.5} \ci{46.0}{53.0} & \underline{57.6} \ci{54.4}{61.0} \\
32B & SFT+DPO & \underline{51.4} \ci{47.8}{55.1} & \textbf{61.6} \ci{58.4}{64.9} & \textbf{52.2} \ci{48.8}{55.9} & \textbf{59.8} \ci{56.4}{63.2} \\
\bottomrule
\end{tabular}

%% file: tables/visual_audit.tex
\begin{tabular}{llccccc}
\toprule
& & \multicolumn{3}{c}{All requests} & \multicolumn{2}{c}{Long requests} \\
\cmidrule(lr){3-5} \cmidrule(lr){6-7}
Size & Method & Precision & Recall & Unsupported & Precision & Unsupported \\
\midrule
4B & Prompt & 95.2 & 29.7 & 20.8 & 94.1 & 27.5 \\
 & SFT & 96.2 & 31.4 & 14.6 & 96.7 & 15.0 \\
 & SFT+DPO & 96.0 & 31.5 & 16.7 & 96.7 & 15.0 \\
\bottomrule
\end{tabular}

%% file: tables/oracle_latency.tex
\begin{tabular}{lccc}
\toprule
4B method & Strict (\%) & Diff.\ (pp) & Median s \\
\midrule
Prompting & 24.5 & $-23.6$ \ci{-27.6}{-19.6} & -- \\
Best-of-two & 33.8 & $-14.4$ \ci{-18.6}{-10.1} & 1.69 \\
Best-of-four & 45.1 & $-3.0$ \ci{-7.5}{+1.5} & 2.26 \\
SFT (unmerged) & 48.1 & -- & 1.69 \\
SFT (merged) & 47.0 & $-1.1$ \ci{-2.6}{+0.3} & 1.36 \\
\bottomrule
\end{tabular}

%% file: tables/diversity.tex
\begin{tabular}{llccccc}
\toprule
Size & Method & Distinct-1 & Distinct-2 & Opening & Jaccard & Rep.\ bigram \\
\midrule
4B & Prompt & 0.391 & 0.778 & 0.033 & 0.076 & 0.029 \\
 & SFT & 0.357 & 0.742 & 0.030 & 0.091 & 0.029 \\
 & SFT+DPO & 0.351 & 0.732 & 0.043 & 0.096 & 0.035 \\
\midrule
8B & Prompt & 0.355 & 0.749 & 0.025 & 0.088 & 0.038 \\
 & SFT & 0.353 & 0.741 & 0.031 & 0.093 & 0.029 \\
 & SFT+DPO & 0.356 & 0.736 & 0.049 & 0.096 & 0.031 \\
\midrule
32B & Prompt & 0.335 & 0.728 & 0.030 & 0.100 & 0.075 \\
 & SFT & 0.351 & 0.737 & 0.029 & 0.093 & 0.029 \\
 & SFT+DPO & 0.348 & 0.734 & 0.034 & 0.097 & 0.030 \\
\bottomrule
\end{tabular}

%% file: 11_transfer_rules.tex
\section{Changed-instruction Evaluation Rules}
\label{app:transfer-rules}

The changed-instruction evaluation uses 100 new COCO images and eight combinations of vocabulary, syntax, and length requests.
The prompt and rules were fixed using 20 separate development images before final generation.
The evaluated students were trained on the earlier 500 training images using only the original instructions.
We use the original experiment's target names: basic and advanced for vocabulary, and simple and complex for syntax. The saved experiment records use A2-like for basic vocabulary and simple syntax, and B1-like for advanced vocabulary and complex syntax. The prompts and scoring rules below differ from those in the original evaluation.

\paragraph{Paired training differences.}
Under changed instructions, SFT+DPO improves three-axis success over SFT by 7.00 pp [$+4.00$, $+10.00$] at 4B and 5.38 pp [$+2.50$, $+8.12$] at 8B.
Brackets give 95\% paired image-bootstrap intervals; each comparison uses the same 100 images and 800 outputs per method.

\subsection{Inference setup}

We ran student inference on a Mac with merged adapters and batches of 16 at 4B/8B and eight at 32B; their original evaluations used unmerged adapters on Linux.
Batch 16 also differs from the pilot's batch eight, and we have not isolated the numerical effects of these changes.

\subsection{Example outputs}
\label{app:transfer-examples}

Table~\ref{tab:transfer-example} shows four outputs for the image in Figure~\ref{fig:transfer-example}, with one requested attribute changed at a time.

\begin{table*}[t]
  \centering
  \small
  \setlength{\tabcolsep}{3pt}
  \input{tables/followup_examples}
  \caption{Selected, unedited 32B SFT+DPO outputs for the source image in Figure~\ref{fig:transfer-example}, the first image listed in the changed-instruction evaluation (COCO 52017). Targets are listed as vocabulary, syntax, and total words. Each row changes one request from the baseline. All four pass vocabulary, syntax, and format, but only the long request passes length. The syntax row contains ``which has the code SP-AWF.'' These examples illustrate length failures without estimating their frequency. Image faithfulness is unverified: the photograph does not establish ``flying high''.}
  \label{tab:transfer-example}
\end{table*}

\subsection{Generation prompts}

All direct-generation methods use the following template. The fields contain the full instructions below.

\begin{quote}\small
Describe this image in English.

Vocabulary: \{vocabulary\_instruction\}\newline
Grammar: \{grammar\_instruction\}\newline
Length: Write \{length\_range\} words in total, in \{sentence\_range\} sentences.

Write complete sentences. Describe only what is visible in the image. Do not repeat words or ideas to reach the length.\newline
Return only the description.
\end{quote}

\paragraph{Vocabulary fields.}
The basic instruction is ``Use only basic, everyday words that a learner at about CEFR A2 level knows. Avoid less common or advanced words.''
The advanced instruction is ``Use mostly common words, but about one in five content words (and at least two) should be B1-level words (intermediate, less common everyday words) that fit the image. Avoid rare or advanced words.''

\paragraph{Grammar fields.}
The simple instruction is ``Use only simple sentences. You may join two parts with and, but, or so. Do not use relative clauses (who, which, that) or subordinate clauses (because, when, while, if, although, as), and do not begin with an -ing or -ed phrase.''
The complex instruction is ``Include at least one relative clause (with who, which, or that) or subordinate clause (with because, when, while, if, although, or as) that has its own subject and verb. Do not put one such clause inside another.''

\paragraph{Length fields.}
The word range is ``15--20'' with ``one to three'' sentences, or ``30--40'' with ``two to five'' sentences. The actual word ranges use en dashes.

\paragraph{Two-stage prompts.}
The first stage receives the image and ``Describe the image factually in a few sentences.''
The second stage receives no image. It uses the direct-generation template with the opening sentence replaced by:
\begin{quote}\small
Describe the scene in the following description in English.

Description: \{description\}
\end{quote}
Here, \{description\} is the first-stage output. The phrase ``visible in the image'' is replaced by ``stated in the description''; all other instructions are identical.

\subsection{Scoring rules}

\paragraph{Vocabulary.}
The parser identifies nouns, verbs, adjectives, and adverbs and maps their lemmas and parts of speech to the same word lists used in the original evaluation.
Repeated lemma--part-of-speech pairs count once; proper nouns, auxiliary verbs, and numbers are excluded.
Fewer than four recognized types or coverage below 50\% makes the estimate undeterminable, which cannot pass.
An A1--A2 estimate requires at most one B2-or-higher type and either a B1-or-higher share no greater than 20\% or at most one B1-or-higher type.
A B2-or-higher estimate requires at least two B2-or-higher types and a share of at least 25\%.
Other determinable outputs receive the B1 estimate.
The judge estimates vocabulary without access to the requested target. Basic requests pass with an A1--A2 estimate; advanced requests pass with a B1 estimate.
These estimates do not certify the CEFR proficiency level of a complete description.

\begin{figure}[t]
  \centering
  \includegraphics[width=0.72\columnwidth]{followup_052017.jpg}
  \caption{Source image (COCO 52017) for the outputs in Table~\ref{tab:transfer-example}. Photo license recorded by COCO: CC BY-NC-ND 2.0.}
  \label{fig:transfer-example}
\end{figure}

\paragraph{Syntax.}
The original finite-clause detector supplies relative and subordinate clauses.
The simple target requires no counted clause and excludes absolute constructions and sentence-initial participial clauses.
The complex target requires at least one counted clause and excludes multiple embedding and absolute constructions.
Coordination is allowed; passive relatives and trailing participial clauses are recorded but do not alone fail complex syntax.
A finite verb is required for a determinable grammar estimate.

\paragraph{Length and format.}
A word is a run of letters or digits; internal apostrophes and hyphens remain part of a word after Unicode normalization.
Length must fall exactly within 15--20 or 30--40 words across the entire output.
Format requires normal completion without headings, bullets, or prefatory text and one to three sentences for short requests or two to five for long requests.
Three-axis success requires determinable vocabulary and syntax matching their targets and length within range; success with format additionally requires the format check.
Empty, erroneous, and undeterminable outputs remain in the denominator.

\subsection{Object-category mentions}
The category matcher accesses WordNet \citep{miller-1995-wordnet} through NLTK \citep{bird-etal-2009-nltk}.
\label{app:object-mentions}

For each output, we count the distinct annotated COCO object categories mentioned in the text. We sum these counts across outputs and divide by the sum of annotated-category counts for their images, counting each image once per output. The comparison uses the same 98 images with object annotations and 784 outputs per method. Mentions match noun lemmas or two-word lemma sequences against category word lists. For lemmas absent from all lists, the first WordNet noun sense must equal a category synset or have it as a hypernym. This measures recall of category names; it can miss paraphrases, and annotations include small background objects. It does not assess the correctness of statements about those objects or human-rated faithfulness.

\paragraph{Reproducibility and interpretation.}
The nine combinations of student size and method, the teacher reference, and the two-stage pipeline each cover the same 800 image--condition inputs.
The manuscript table builder checks the stored rule decisions and word counts for all 8,800 outputs against the archived summaries and preserves source hashes.
The selected example uses the first final-evaluation image in manifest order and four conditions changing one requested axis at a time.
It is a descriptive illustration; it is neither sampled to estimate prevalence nor a human faithfulness evaluation.

\subsection{Supplementary analyses of 32B length compliance}
\label{app:32b-length}

\paragraph{Length compliance under changed instructions.}
Training improves control under the original instructions, but length compliance remains poor under the changed instructions.
The following analyses use the same distilled students as the main comparison.
On the first original image set, the 32B SFT checkpoint averaged 12.61 and 22.45 words for the 10--14 and 20--24 requests, with 85.00\% strict length compliance.
Under the changed requests, the same checkpoint averages 25.15 and 41.92 words, with 20.88\% compliance.
Switching from short to long requests adds 16.77 words [16.23, 17.34], close to the 17.5-word difference between target midpoints.
The model thus responds to the requested increase but exceeds the absolute ranges; the switch also changes the permitted sentence count.
Because wording, targets, sentence counts, images, and inference environment change together, we cannot attribute the compliance gap to any one factor.

\paragraph{Cross-axis effects under changed instructions.}
For each image and length range in the changed-instruction evaluation, we compare vocabulary requests while holding syntax fixed, or syntax requests while holding vocabulary fixed, then average across the two settings of the unchanged axis.
At 32B SFT, changing vocabulary from basic to advanced adds 2.69 words under long requests, and changing syntax from simple to complex adds 1.78 words (Table~\ref{tab:interference}).
Both paired image-bootstrap intervals exclude zero; the corresponding short-request differences are uncertain.
Relative to the basic-vocabulary and simple-syntax baseline, requesting advanced vocabulary and complex syntax together adds fewer words than the sum of changing each alone: the interaction, averaged across length ranges, is negative ($-1.11$ words [$-2.07$, $-0.12$]).
The difference in this interaction between long and short requests is $-0.05$ words and is uncertain.
These exploratory intervals are unadjusted for multiple comparisons.

\begin{table}[t]
  \centering
  \small
  \setlength{\tabcolsep}{1.5pt}
  \input{tables/followup_interference}
  \caption{32B word-count changes when vocabulary changes from basic to advanced or syntax from simple to complex. Each cell averages two comparisons per image (100 images, 200 pairs), one at each setting of the other axis. That setting and length range are fixed within each pair. Brackets are 95\% image-bootstrap intervals.}
  \label{tab:interference}
\end{table}

Excess length persists under the easier requests: 190 of 200 short, simple-syntax outputs from 32B SFT exceed 20 words.
With basic vocabulary and simple syntax, short outputs average 24.85 words and only 7/100 pass length.
Long SFT outputs have similar mean sentence counts at 32B and 4B (4.33 and 4.39), but the mean of per-output words per sentence is 10.04 versus 7.77.
Even among short 32B SFT outputs with one or two sentences, 126/142 exceed 20 words.
Neither harder vocabulary/syntax requests nor sentence counts alone explain the excess length.

\paragraph{Length failures precede preference training.}
Before task-specific training, the 32B model already exceeds the changed ranges in 744/800 outputs.
This student's 426-pair dataset contains 71 too-short and 71 too-long length rejections. Across all 426 pairs, preferred responses average 0.056 more words than rejected responses.

\paragraph{Limits of indirect length supervision.}
Text-only evaluations also find difficulties with fine-grained constraints, including word-count planning \citep{sun-etal-2023-evaluating}.
SFT learns token prediction from compliant targets; DPO learns preferences over complete responses. Neither imposes a hard word limit during generation.
Continuing to describe visible details or complete a requested clause could exceed the range, but we have not shown that this explains the size difference.
The saved 32B outputs all terminate normally, and the original and changed-instruction word counters disagree for only eight of 800 outputs for each of the SFT and SFT+DPO methods. Truncation and counting conventions therefore cannot explain the widespread excess length.
Among the 253 long-request SFT outputs above 40 words, 233 (92.1\%) contain an earlier boundary detected by the sentence-counting rule within 30--40 words, after which generation continues.
Stopping at those boundaries could remove required vocabulary, a clause, or relevant content, or violate the sentence-count requirement.
Testing this possibility requires end-token probabilities or a comparison of decoding with an explicit word budget on the same inputs, checking all constraints and content.

\paragraph{Evidence needed for a causal explanation.}
A controlled follow-up should vary sentence-count instructions, original versus changed word ranges, and vocabulary/syntax wording separately, holding images, weights, precision, and decoding fixed.
Training a 32B student on the changed format would test whether training on those instructions reduces the transfer gap.

%% file: tables/followup_examples.tex
\begin{tabular}{@{}p{0.22\textwidth}p{0.62\textwidth}rr@{}}
\toprule
Changed request & Unedited output & Words & Length \\
\midrule
Baseline (basic, simple, 15--20) & A yellow and red airplane flies in the sky. It has black letters SP-AWF on its side. The plane has a blue front and two wheels under it. & 28 & Fail \\
Vocabulary (advanced, simple, 15--20) & A yellow and red airplane flies through a gray sky. Its registration number SP-AWF is clearly visible on the side. The propeller spins at the front. & 26 & Fail \\
Syntax (basic, complex, 15--20) & A yellow and red plane flies in the sky. It has a blue front and black wheels. The plane, which has the code SP-AWF, is flying high. & 27 & Fail \\
Length (basic, simple, 30--40) & A small yellow and red airplane flies in the sky. The plane has black letters "SP-AWF" on its side. It has a blue front part and a white propeller. The wheels are down, and the sky is gray. & 38 & Pass \\
\bottomrule
\end{tabular}

%% file: tables/followup_interference.tex
\begin{tabular}{lrr}
\toprule
Changed axis & 15--20 words & 30--40 words \\
\midrule
SFT, Vocab. & +0.13 \ci{-0.36}{+0.62} & +2.69 \ci{+1.73}{+3.63} \\
SFT, Syntax & -0.07 \ci{-0.70}{+0.61} & +1.78 \ci{+0.98}{+2.59} \\
SFT+DPO, Vocab. & +0.60 \ci{+0.10}{+1.10} & +2.93 \ci{+2.05}{+3.77} \\
SFT+DPO, Syntax & +0.08 \ci{-0.51}{+0.68} & +2.94 \ci{+2.15}{+3.71} \\
\bottomrule
\end{tabular}

%% file: 12_cross_axis_details.tex
\section{Cross-axis Analysis Under the Original Instructions}
\label{app:cross-axis}

This analysis uses the original eight conditions and the same two image sets as Table~\ref{tab:main}.
The separate 32B analysis under changed instructions appears in Appendix~\ref{app:32b-length}.
Tables~\ref{tab:cross-axis-fixed} and~\ref{tab:cross-axis-numeric} report every overall fixed-request and numerical comparison for the final 4B SFT+DPO checkpoint.
Each contrast contains 400 pairs from 100 images per set; all four differences for an image receive equal weight.
The difference is advanced minus basic, complex minus simple, or long minus short, according to the changed request.
A negative compliance difference means that fewer outputs satisfy the fixed request after the change.
Both passing and failing outputs enter each contrast; the mean difference is not a failure probability conditional on initially passing.

\paragraph{Estimation and inference.}
The analysis uses NumPy 2.4.6 \citep{harris-etal-2020-numpy} and SciPy 1.17.1 \citep{virtanen-etal-2020-scipy}.
Let $d_i$ be the mean of the four paired differences for image $i$.
We test whether the population mean of $d_i$ is zero using a two-sided one-sample $t$ test with 99 degrees of freedom, equivalent to a paired test on image-level means.
Paired Cohen's $d_z$ is $\bar d/s_d$, where $s_d$ is the sample standard deviation of the 100 image differences.
It standardizes differences across images, rather than across individual descriptions.
Percentile intervals for both $\bar d$ and $d_z$ use 10,000 whole-image bootstrap resamples, shared across contrasts within each set (seeds 20261012 and 20261013).
All 46 reported contrasts have 100 complete image differences and 10,000 defined effect-size resamples.
The intervals are pointwise and remain unadjusted for multiple comparisons.

\paragraph{Exploratory correction families.}
The broader analysis covers 4B, 8B, and 32B at Prompt, SFT, and SFT+DPO, using 14,400 existing outputs.
After inspecting effects and intervals, we specified five families before calculating $p$-values: 16 final-4B fixed-request tests; 128 fixed-request tests for the other eight size/stage combinations; 270 numerical tests; 84 matched training-stage compliance tests; and 1,656 conditional tests.
The numerical family tests five indicators (B1+ share, clause count, clause presence, word count, and lexical coverage) for three changed axes, nine size/stage combinations, and two sets: $5\times3\times9\times2=270$.
The stage family compares Prompt with SFT and SFT with SFT+DPO at each of three sizes on both sets, for seven compliance indicators: vocabulary, syntax, format, strict and tolerant length, and strict and tolerant joint success ($2\times3\times2\times7=84$).
Each stage contrast averages eight matched condition differences within an image.
The conditional family repeats the 144 fixed-request and 270 numerical tests separately for each of the four combinations of the other two requests, instead of averaging over those combinations: $(144+270)\times4=1656$.
Holm correction uses $\alpha=.05$ within each family.
The 498-test sensitivity correction combines the first four families, which average over all relevant condition pairs within an image; it excludes the 1,656 conditional tests.
Thus, $p_{498}$ does not correct across every exploratory analysis.
This retrospective choice of families limits confirmatory interpretation even when an adjusted $p$-value is below .05.

\paragraph{Sensitivity and transitions.}
Five of the 16 final-4B fixed-request tests reject after the 16-test Holm correction: length-to-syntax and syntax-to-strict-length on both sets, and vocabulary-to-tolerant-length on set 1.
None of these fixed-request tests rejects after the broader 498-test correction.
Changing vocabulary from basic to advanced reduces tolerant length compliance by 4.25 and 2.75 pp on sets 1 and 2, respectively; only the first reduction survives the 16-test correction.
Changing the length request from short to long increases the B1-or-higher word share by 3.61 and 5.14 pp, which survives the 498-test correction ($p=.0231$ and $p<.00001$).
This numerical change does not itself imply failure of the vocabulary request, whose pass criteria depend on the requested level.

The $t$ test assumes independent images and an adequate approximation for the distribution of image-mean differences.
A sensitivity analysis for the 498 overall tests flips the signs of whole-image differences using 100,000 shared sign assignments per set (seeds 2026101301 and 2026101302).
The two-sided Monte Carlo $p$-value is $(b+1)/(100000+1)$, where $b$ counts assignments with an absolute mean at least as large as observed.
This assumes sign symmetry under the null; it does not reflect randomized assignment of instructions.
All five fixed-request findings that survive the 16-test $t$-test correction also survive the corresponding sign-flip correction.
Pass-to-fail and fail-to-pass counts in Table~\ref{tab:cross-axis-fixed} describe the 400 pairs and can offset one another in the image-mean test.
For example, changing syntax produces 51 and 57 vocabulary pass-to-fail transitions, alongside 43 and 45 fail-to-pass transitions, on sets 1 and 2, respectively.
Neither non-rejection nor small net change establishes equivalence or preservation of individual outputs.

\begin{table*}[t]
  \centering
  \footnotesize
  \setlength{\tabcolsep}{3pt}
  \input{tables/cross_axis_fixed}
  \caption{All fixed-request comparisons for final 4B SFT+DPO under the original instructions. Set 1 and set 2 each contain 100 images and 400 pairs per operation. Changes are basic to advanced vocabulary, simple to complex syntax, or short to long length; the other two requests remain fixed within each pair. $\Delta$ is the change in compliance in percentage points. Brackets give pointwise 95\% image-bootstrap intervals. $p_{16}$ and $p_{498}$ are Holm-adjusted two-sided $t$-test values for the 16-test family and broader sensitivity family. PF / FP count pass-to-fail / fail-to-pass transitions. Strict and $\pm2$ refer to exact and two-word-tolerant length scoring.}
  \label{tab:cross-axis-fixed}
\end{table*}

\begin{table*}[t]
  \centering
  \footnotesize
  \setlength{\tabcolsep}{4pt}
  \input{tables/cross_axis_numeric}
  \caption{All numerical comparisons for final 4B SFT+DPO under the original instructions, using the same pairs as Table~\ref{tab:cross-axis-fixed}. $\Delta$ is the after-minus-before mean change. B1+ share is the proportion of recognized content-word occurrences at B1 or above; lexical coverage is the recognized share of all content-word occurrences. Clause presence is a binary indicator of at least one detected clause. Proportion changes use percentage points (pp); clause and word counts use their original units. Brackets are pointwise 95\% image-bootstrap intervals; $p_{498}$ is Holm-adjusted across the 498 overall exploratory tests. A numerical change can occur while the fixed request still passes.}
  \label{tab:cross-axis-numeric}
\end{table*}

%% file: tables/cross_axis_fixed.tex
\begin{tabular}{lllrrrrr}
\toprule
Set & Changed request & Fixed target & $\Delta$ pp [95\% CI] & $d_z$ [95\% CI] & $p_{16}$ & $p_{498}$ & PF / FP \\
\midrule
1 & Vocabulary & Syntax & $-0.25$ [-2.25, +1.50] & $-0.026$ [-0.207, +0.175] & 1.0000 & 1.0000 & 7 / 6 \\
1 & Vocabulary & Length ($\pm2$) & $-4.25$ [-7.00, -1.75] & $-0.308$ [-0.463, -0.141] & 0.0319 & 0.6886 & 24 / 7 \\
1 & Vocabulary & Length (strict) & $-4.25$ [-9.50, +1.25] & $-0.158$ [-0.360, +0.044] & 1.0000 & 1.0000 & 69 / 52 \\
1 & Syntax & Vocabulary & $-2.00$ [-7.00, +3.00] & $-0.077$ [-0.281, +0.113] & 1.0000 & 1.0000 & 51 / 43 \\
1 & Syntax & Length ($\pm2$) & $-1.75$ [-4.25, +1.00] & $-0.130$ [-0.346, +0.065] & 1.0000 & 1.0000 & 18 / 11 \\
1 & Syntax & Length (strict) & $-11.75$ [-18.25, -5.50] & $-0.354$ [-0.578, -0.160] & 0.0087 & 0.1729 & 96 / 49 \\
1 & Length & Vocabulary & $+2.00$ [-2.50, +6.50] & $+0.087$ [-0.113, +0.287] & 1.0000 & 1.0000 & 40 / 48 \\
1 & Length & Syntax & $-4.75$ [-7.50, -2.25] & $-0.361$ [-0.478, -0.228] & 0.0072 & 0.1360 & 20 / 1 \\
\midrule
2 & Vocabulary & Syntax & $+1.50$ [-0.50, +3.50] & $+0.151$ [-0.050, +0.319] & 1.0000 & 1.0000 & 4 / 10 \\
2 & Vocabulary & Length ($\pm2$) & $-2.75$ [-6.00, +0.50] & $-0.165$ [-0.372, +0.030] & 1.0000 & 1.0000 & 26 / 15 \\
2 & Vocabulary & Length (strict) & $-4.00$ [-9.00, +0.75] & $-0.160$ [-0.352, +0.032] & 1.0000 & 1.0000 & 67 / 51 \\
2 & Syntax & Vocabulary & $-3.00$ [-8.25, +2.25] & $-0.115$ [-0.326, +0.082] & 1.0000 & 1.0000 & 57 / 45 \\
2 & Syntax & Length ($\pm2$) & $-2.25$ [-5.25, +0.75] & $-0.149$ [-0.363, +0.047] & 1.0000 & 1.0000 & 24 / 15 \\
2 & Syntax & Length (strict) & $-10.50$ [-16.75, -4.00] & $-0.318$ [-0.544, -0.121] & 0.0253 & 0.5103 & 96 / 54 \\
2 & Length & Vocabulary & $+6.00$ [+1.00, +11.25] & $+0.234$ [+0.045, +0.430] & 0.2349 & 1.0000 & 43 / 67 \\
2 & Length & Syntax & $-5.00$ [-7.75, -2.50] & $-0.376$ [-0.500, -0.248] & 0.0046 & 0.0824 & 21 / 1 \\
\bottomrule
\end{tabular}

%% file: tables/cross_axis_numeric.tex
\begin{tabular}{lllrrr}
\toprule
Set & Changed request & Indicator & $\Delta$ [95\% CI] & $d_z$ [95\% CI] & $p_{498}$ \\
\midrule
1 & Vocabulary & B1+ share (pp) & $+19.55$ [+17.29, +21.83] & $+1.652$ [+1.435, +1.957] & $<.0001$ \\
1 & Vocabulary & Clause count & $-0.01$ [-0.05, +0.04] & $-0.022$ [-0.230, +0.171] & 1.0000 \\
1 & Vocabulary & Clause presence (pp) & $+0.25$ [-1.50, +2.25] & $+0.026$ [-0.175, +0.207] & 1.0000 \\
1 & Vocabulary & Word count & $+0.34$ [+0.10, +0.57] & $+0.280$ [+0.081, +0.483] & 1.0000 \\
1 & Vocabulary & Lexical coverage (pp) & $-10.49$ [-12.03, -9.01] & $-1.332$ [-1.586, -1.146] & $<.0001$ \\
1 & Syntax & B1+ share (pp) & $+0.40$ [-1.16, +1.91] & $+0.050$ [-0.142, +0.258] & 1.0000 \\
1 & Syntax & Clause count & $+1.26$ [+1.20, +1.33] & $+3.715$ [+2.964, +4.888] & $<.0001$ \\
1 & Syntax & Clause presence (pp) & $+93.25$ [+89.75, +96.25] & $+5.614$ [+4.338, +8.252] & $<.0001$ \\
1 & Syntax & Word count & $+1.00$ [+0.73, +1.28] & $+0.714$ [+0.510, +0.963] & $<.0001$ \\
1 & Syntax & Lexical coverage (pp) & $+0.58$ [-0.48, +1.64] & $+0.104$ [-0.087, +0.301] & 1.0000 \\
1 & Length & B1+ share (pp) & $+3.61$ [+1.89, +5.33] & $+0.413$ [+0.206, +0.670] & 0.0231 \\
1 & Length & Clause count & $+0.27$ [+0.22, +0.32] & $+1.018$ [+0.844, +1.232] & $<.0001$ \\
1 & Length & Clause presence (pp) & $+4.75$ [+2.25, +7.50] & $+0.361$ [+0.228, +0.478] & 0.1360 \\
1 & Length & Word count & $+9.94$ [+9.68, +10.20] & $+7.428$ [+6.781, +8.349] & $<.0001$ \\
1 & Length & Lexical coverage (pp) & $-0.59$ [-1.63, +0.48] & $-0.108$ [-0.323, +0.085] & 1.0000 \\
\midrule
2 & Vocabulary & B1+ share (pp) & $+17.54$ [+15.52, +19.61] & $+1.670$ [+1.415, +2.030] & $<.0001$ \\
2 & Vocabulary & Clause count & $-0.04$ [-0.09, -0.00] & $-0.197$ [-0.382, -0.011] & 1.0000 \\
2 & Vocabulary & Clause presence (pp) & $-1.50$ [-3.50, +0.50] & $-0.151$ [-0.319, +0.050] & 1.0000 \\
2 & Vocabulary & Word count & $+0.33$ [+0.10, +0.55] & $+0.289$ [+0.089, +0.507] & 1.0000 \\
2 & Vocabulary & Lexical coverage (pp) & $-9.62$ [-11.19, -8.11] & $-1.218$ [-1.446, -1.041] & $<.0001$ \\
2 & Syntax & B1+ share (pp) & $+0.82$ [-0.56, +2.30] & $+0.111$ [-0.082, +0.302] & 1.0000 \\
2 & Syntax & Clause count & $+1.32$ [+1.28, +1.36] & $+5.942$ [+5.394, +6.719] & $<.0001$ \\
2 & Syntax & Clause presence (pp) & $+93.00$ [+89.50, +96.00] & $+5.570$ [+4.308, +8.104] & $<.0001$ \\
2 & Syntax & Word count & $+0.95$ [+0.68, +1.21] & $+0.693$ [+0.467, +0.974] & $<.0001$ \\
2 & Syntax & Lexical coverage (pp) & $+1.51$ [+0.52, +2.55] & $+0.286$ [+0.104, +0.466] & 1.0000 \\
2 & Length & B1+ share (pp) & $+5.14$ [+3.51, +6.75] & $+0.619$ [+0.413, +0.868] & $<.0001$ \\
2 & Length & Clause count & $+0.32$ [+0.27, +0.37] & $+1.211$ [+1.007, +1.493] & $<.0001$ \\
2 & Length & Clause presence (pp) & $+5.00$ [+2.50, +7.75] & $+0.376$ [+0.248, +0.500] & 0.0824 \\
2 & Length & Word count & $+10.09$ [+9.82, +10.37] & $+7.097$ [+6.219, +8.354] & $<.0001$ \\
2 & Length & Lexical coverage (pp) & $-1.13$ [-2.21, -0.08] & $-0.207$ [-0.403, -0.015] & 1.0000 \\
\bottomrule
\end{tabular}

%% file: 13_human_evaluation_details.tex
\section{Human Evaluation Details}
\label{app:human}

\paragraph{Participants and procedure.}
The analyzed sample comprises 24 Prolific participants, including three pilot participants and 21 subsequently recruited participants.
Participants were randomly sampled from eligible English speakers on Prolific.
Their ages range from 27 to 61 years ($M=40.13$, $SD=11.21$); the demographic export records 11 female and 13 male participants, all with English listed as their language.
They reside in the United Kingdom (9), Canada (6), United States (5), South Africa (2), Mexico (1), and Romania (1).
The study recruited native English speakers to judge difficulty for an English learner.

We randomly sampled 20 of the second evaluation set's 100 images without replacement, using seed 20261012, without selecting on automatic scores or human ratings.
We used the final 4B SFT+DPO model's unedited outputs for all eight vocabulary/syntax/length combinations on these images.
The 160 distinct descriptions each receive three evaluations.
Each of eight lists contains the same 20 images with one condition per image; across lists, every image appears under all eight conditions.
Each list contains ten descriptions per level of each axis and two or three per condition combination.
Each participant evaluates one list, and each list is evaluated by three participants.
The three pilot participants evaluate the first list; the 21 subsequently recruited participants evaluate the other seven.
The records establish this allocation but not random assignment of participants to lists.
Each list was shuffled with a fixed seed; its three raters see the same order, without further participant-specific randomization.
Each description and its corresponding photograph appear on the same page, followed by the five rating questions; smaller screens may require scrolling.
The model identity, requested conditions, and automatic scores are hidden from raters.

\paragraph{Instructions and practice.}
Participants were instructed to judge vocabulary and sentence-structure difficulty separately from the perspective of ``someone learning English,'' with no specific proficiency level prescribed.
Clause presence was assessed separately from perceived difficulty.
Before rating, participants received clause definitions and four practice questions illustrating counted clauses and excluded constructions, followed by correct answers, explanations, and a required acknowledgment.
The instructions explicitly stated that practice responses would not be used to reject submissions.
Participants were instructed to avoid AI and grammar checkers and to select ``Cannot judge'' when uncertain.

\paragraph{Rating questions.}
The vocabulary question is:
\begin{quote}
How difficult are the WORDS themselves for someone learning English? Focus on vocabulary, ignoring sentence structure and length as far as possible.
\end{quote}
The syntax question is:
\begin{quote}
How difficult is the SENTENCE STRUCTURE for someone learning English? Imagine all words are familiar. Focus on grammatical connections and embedding, not vocabulary difficulty.
\end{quote}
Vocabulary and syntax difficulty use 1 (very easy), 2 (easy), 3 (moderately difficult), 4 (difficult), and 5 (very difficult).
Both questions also offer ``Cannot judge''.
The separate clause-presence question is:
\begin{quote}
Does this description contain at least one finite subordinate clause or a relative clause as defined in the guide? Do not count a prepositional phrase, participial phrase, infinitive, or coordination alone.
\end{quote}
Its response options are Yes, No, and Cannot judge.
The unsupported-content question is:
\begin{quote}
Does the description include ANY content that is not supported by the photograph? If the image cannot be viewed, choose Cannot judge.
\end{quote}
Its response options are Yes, No, and Cannot judge.
The naturalness question is:
\begin{quote}
How natural is this description as English? Consider grammar and idiomatic phrasing, separately from image accuracy.
\end{quote}
Its options are 1 (very unnatural), 2 (somewhat unnatural), 3 (neither natural nor unnatural), 4 (somewhat natural), 5 (very natural), and Cannot judge.
``Cannot judge'' is retained as missing for the corresponding outcome only: two vocabulary ratings, one syntax rating, no naturalness ratings, 16 clause judgments, and 14 unsupported-content judgments.

\paragraph{Clause guide and practice materials.}
The clause definition provided to participants was:
\begin{quote}\small
CLAUSE PRESENCE: This is a separate yes/no structure check, not a difficulty score. Count a finite subordinate clause (a dependent unit with a tensed verb, possibly with an auxiliary) or a relative clause introduced by who, which, that, whose, where, when, etc. Count relative clauses even when the relative word is omitted: the bag she carries. Examples: while a child watches; a dog that is sleeping. Do not count a prepositional phrase (near the tree), a participial phrase alone (a man holding a bag), an infinitive alone (to carry a bag), or coordination alone (a man sits and a woman stands). The words that/while/when alone are not sufficient: consider their grammatical role.
\end{quote}
Each of the following four descriptions was followed by ``Does this contain a clause counted by our definition?'' with Yes, No, and Cannot judge as options:
\begin{enumerate}
\item A man holding a red bag stands near a tree.
\item A man who is holding a red bag stands near a tree.
\item A man stands near a tree while a child watches.
\item A man sits and a woman stands beside him.
\end{enumerate}
The next page gave these explanations:
\begin{quote}\small
1: No. holding a red bag is a participial phrase; near a tree is a prepositional phrase. Neither is a counted clause.

2: Yes. who is holding a red bag is a finite relative clause (is is tensed).

3: Yes. while a child watches is a finite subordinate clause.

4: No. and joins two main clauses; coordination alone is not counted.

The presence of a counted clause does not automatically make a description difficult. Give an independent sentence-structure difficulty rating. Uncommon words alone also do not make sentence structure difficult.
\end{quote}
The required acknowledgment stated ``Please review the explanations before continuing. You may refer back to this guide,'' with the response ``I have read the explanations.''

\paragraph{Study information, consent, and compensation.}
The opening study information read:
\begin{quote}\small
In this study you will read 20 English descriptions of 20 different photographs and give five short judgments for each. You will first read a guide and complete four practice questions with explanations. Please complete only ONE set. Your judgments may differ from other readers; there are no expected difficulty ratings. Participation is voluntary. You may stop before submitting. Do not enter your name or email. Contact the researcher through Prolific with any questions.
\end{quote}
The required consent item read:
\begin{quote}\small
I have read the study description, am at least 18, and voluntarily agree to take part. If you do not agree, close this form without submitting.
\end{quote}
All included participants selected ``I agree.''
Prolific IDs were collected with the instruction ``Copy your Prolific ID here. Do not enter your name or email address.''
Compensation was set at \pounds9 per hour, following Prolific's recommended payment level,\footnote{\url{https://researcher-help.prolific.com/en/articles/445266-how-much-should-i-pay-participants}}
and the achieved hourly rate based on actual completion times was also \pounds9.

\paragraph{Response screening.}
Three additional respondents were excluded after inspection of their responses.
We made these exclusions through manual quality review of check-question answers and repetitive response patterns, without a preregistered exclusion rule.
Two instruction checks required selecting Blue and Triangle. All 27 respondents, including those excluded, answered both correctly, so these items alone do not explain the exclusions.
The analysis retains the three pilot participants.
Because further recruitment followed inspection of the pilot and some exclusions were made after inspection, we treat the results as exploratory.

\paragraph{Statistical analysis.}
We treat the 1--5 ratings as numeric outcomes in ordinary least squares models.
Vocabulary, syntax, and length requests are coded $-0.5$ and $+0.5$; the model includes all two-way and three-way interactions and fixed effects for participant and image.
Each main coefficient is the higher-minus-lower request difference, equally averaged over the other two requests.
Descriptive means first average valid ratings within each description, then weight descriptions equally.
The adjusted differences additionally account for participant and image effects and can differ slightly from these descriptive differences.

We use two-way cluster-robust covariance estimates, clustering by participant and image and subtracting their intersection term.
Each component has a finite-sample CR1 correction.
Intervals and two-sided tests use a $t$ distribution with 19 degrees of freedom, the smaller cluster count minus one.
This is approximate inference with 24 participant clusters and 20 image clusters and a linear approximation to ordinal responses.
The intervals are pointwise.
Holm correction covers the two intended difficulty contrasts together; the seven remaining main effects form a separate family, and the 12 interaction tests form another.
No interaction rejects after the 12-test correction.

\begin{table}[t]
  \centering
  \small
  \setlength{\tabcolsep}{2pt}
  \input{tables/human_secondary}
  \caption{Supplementary human-rating contrasts. Request changes are basic $\to$ advanced vocabulary, simple $\to$ complex syntax, and short $\to$ long length. Differences are adjusted rating points with pointwise 95\% intervals; $p$ values use Holm correction across all seven rows. Valid ratings: vocabulary 478, syntax 479, naturalness 480; all use 24 participants and 20 images.}
  \label{tab:human-secondary}
\end{table}

\paragraph{Cross-axis effects and description quality.}
The vocabulary request increases syntax ratings, and the length request increases both difficulty ratings (Table~\ref{tab:human-secondary}).
These three effects remain positive after the seven-test correction.
The syntax request's estimated effect on vocabulary ratings is uncertain; this does not establish independence.
Mean naturalness is 3.92 across 480 ratings.
None of the three request effects on naturalness rejects after correction, which does not establish equivalent naturalness across conditions.
Raters flag unsupported content in 102 of 466 valid judgments (21.9\%); these are individual judgments, not a consensus error rate over descriptions.
The study evaluates one trained checkpoint and includes no prompt-only or SFT comparison, so it cannot establish that training preserves quality or improves perceived control over those baselines.

\paragraph{Agreement of the automatic clause judge.}
The clause-presence answers above also validate the automatic clause detector used in scoring. Across 464 valid individual judgments (24 raters, 20 images), the detector agrees with the human answer on 68.5\% of judgments (Cohen's $\kappa = 0.37$ [0.28, 0.46]). Aggregating each sentence's judgments by majority vote (152 sentences, ties dropped), agreement rises to 75.0\% ($\kappa = 0.50$ [0.35, 0.64]). The detector is therefore a reasonable but imperfect proxy for the human clause judgments, and pass rates based on it should be read with that error rate in mind.

\paragraph{Blinded quality-rating study.}
The study of Section~\ref{sec:human-quality} used the 20 confirmation-set images whose outputs the three 4B methods generated under all eight requests (480 sentences). Sentences were divided into 36 blinded lists of 40, balanced so that each list mixes methods, images, and requests. Fifteen lists were completed by 16 respondents (one list received two form responses; both are analysed, and one was paid through the platform). Each list interleaved three instruction checks after items 10, 20, and 30 (selecting a named colour or shape); the prespecified rule excluded respondents failing two or more checks, and none did. All 16 respondents self-reported English as their first language and completed the study on a desktop computer. Ratings were analysed with the prespecified image-level paired bootstrap over the 20 images (10,000 resamples); the primary contrasts, fixed before data collection, were SFT minus prompting and SFT+DPO minus SFT on visual faithfulness and naturalness. A sensitivity analysis that centres each respondent's ratings on their own per-dimension mean leaves the conclusions unchanged.

%% file: tables/human_secondary.tex
\begin{tabular}{@{}llcr@{}}
\toprule
Request & Rating & Difference & $p$ \\
\midrule
Syntax & Vocabulary & +0.11 [-0.01, +0.23] & 0.28349 \\
Length & Vocabulary & +0.73 [+0.48, +0.97] & 0.00004 \\
Vocabulary & Syntax & +0.27 [+0.10, +0.44] & 0.01968 \\
Length & Syntax & +0.85 [+0.56, +1.15] & 0.00006 \\
Vocabulary & Naturalness & -0.10 [-0.33, +0.13] & 0.76000 \\
Syntax & Naturalness & -0.05 [-0.24, +0.14] & 0.76000 \\
Length & Naturalness & -0.25 [-0.52, +0.02] & 0.28349 \\
\bottomrule
\end{tabular}